%% file: iclr2026_conference.tex
\documentclass{article} 
\usepackage{iclr2026_conference,times}
\usepackage{xspace}
\usepackage[utf8]{inputenc} 
\usepackage[T1]{fontenc}    
\usepackage{hyperref}       
\usepackage{url}            
\usepackage{booktabs}       
\usepackage{amsfonts}       
\usepackage{nicefrac}       
\usepackage{microtype}      
\usepackage[table]{xcolor}
\usepackage{quoting}
\quotingsetup{vskip=2pt}
\usepackage{makecell}

\usepackage{algorithm}
\usepackage{algpseudocode}
\usepackage{amsmath}
\usepackage{amssymb}
\usepackage{mathtools}
\usepackage{amsthm}
\usepackage{xspace}
\usepackage{enumitem}
\usepackage{tcolorbox}
\usepackage{listings}
\usepackage{wrapfig}
\usepackage{subcaption}
\usepackage{titlesec}
\usepackage{adjustbox}
\hypersetup{colorlinks=true, citecolor=brown, linkcolor=blue}

\usepackage{booktabs,makecell,array}
\newcolumntype{L}[1]{>{\raggedright\arraybackslash}p{#1}}

\definecolor{lightlightgray}{rgb}{0.9, 0.9, 0.9}

\input{math_commands.tex}

\usepackage{hyperref}
\usepackage{url}

\def\method{WorldGym\xspace}

\title{\method: World Model as An Environment for Policy Evaluation}

\author{Julian Quevedo$^{1}$\enskip\enskip
Ansh Kumar Sharma$^{2}$\enskip\enskip
Yixiang Sun$^{2}$\enskip\enskip
Varad Suryavanshi$^{2}$\\
{\bf Percy Liang$^{1}$\enskip\enskip
Sherry Yang$^{1,2,3}$} \\
$^1$Stanford University\enskip\enskip$^2$NYU\enskip\enskip\enskip$^3$Google DeepMind\\
\texttt{julianq@stanford.edu, sherryyang@nyu.edu}
}

\iclrfinalcopy 
\begin{document}

\maketitle

\begin{abstract}
Evaluating robot control policies is difficult: real-world testing is costly, and handcrafted simulators require manual effort to improve in realism and generality. We propose a world-model-based policy evaluation environment (\method), an autoregressive, action-conditioned video generation model which serves as a proxy to real world environments. Policies are evaluated via Monte Carlo rollouts in the world model, with a vision-language model providing rewards. We evaluate a set of VLA-based real-robot policies in the world model using only initial frames from real robots, and show that policy success rates within the world model highly correlate with real-world success rates. Moreoever, we show that \method is able to preserve relative policy rankings across different policy versions, sizes, and training checkpoints. Due to requiring only a single start frame as input, the world model further enables efficient evaluation of robot policies' generalization ability on novel tasks and environments. We find that modern VLA-based robot policies still struggle to distinguish object shapes and can become distracted by adversarial facades of objects. While generating highly realistic object interaction remains challenging, \method faithfully emulates robot motions and offers a practical starting point for safe and reproducible policy evaluation before deployment.\footnote{See videos and code at \href{https://world-model-eval.github.io}{https://world-model-eval.github.io}}

\end{abstract}

\setlength{\abovedisplayskip}{1pt}
\setlength{\abovedisplayshortskip}{1pt}
\setlength{\belowdisplayskip}{1pt}
\setlength{\belowdisplayshortskip}{1pt}
\setlength{\jot}{1pt}

\setlength{\parskip}{0.28em}
\titlespacing\section{0pt}{3pt plus 1pt minus 2pt}{2pt plus 1pt minus 2pt}
\titlespacing\subsection{0pt}{3pt plus 1pt minus 2pt}{2pt plus 1pt minus 2pt}
\makeatletter
\renewcommand{\paragraph}{%
  \@startsection{paragraph}{4}%
  {\z@}{0.05ex \@plus .05ex \@minus .05ex}{-1em}%
  {\normalfont\normalsize\bfseries}%
}

\setlength{\floatsep}{1ex}
\setlength{\textfloatsep}{1ex}
\setlength{\abovecaptionskip}{1ex}
\setlength{\intextsep}{1ex}

\input{introduction}
\input{prelimiaries}
\input{method}

\input{experiment}

\input{related}
\input{conclusion}

\subsubsection*{Acknowledgments}
We thank Xinchen Yan and Doina Precup for reviewing versions of this manuscript. We thank Moo Jin Kim for help in setting up the OpenVLA policy. We thank Boyuan Chen and Kiwhan Song for the Diffusion Forcing GitHub repository.

\bibliography{iclr2026_conference}
\bibliographystyle{iclr2026_conference}
\newpage
\appendix

\input{appendix}

\end{document}

%% file: math_commands.tex
\usepackage{amsmath,amsfonts,bm}

\def\eqref#1{equation~\ref{#1}}

\def\1{\bm{1}}

\DeclareMathAlphabet{\mathsfit}{\encodingdefault}{\sfdefault}{m}{sl}
\SetMathAlphabet{\mathsfit}{bold}{\encodingdefault}{\sfdefault}{bx}{n}



%% file: introduction.tex
\section{Introduction}
\label{sec:intro}



Robots can help humans in ways that range from home robots performing chores~\citep{shafiullah2023bringing,liu2024ok} to hospital robots taking care of patients~\citep{soljacic2024robots}. One of the major road blocks in the development robots lies in evaluation --- how should we ensure that these robots will work reliably without causing any physical damage when deployed in the real world? Traditionally, people have used \emph{handcrafted} software simulators to develop and evaluate robot control policies~\citep{tedrake2019drake,todorov2012mujoco,erez2015simulation}. However, handcrafted simulation based on our understanding of the physical world can be limited, especially when it comes to hardcoding complex dynamics with high degrees of freedom or complex interactions such as manipulating soft objects~\citep{sunderhauf2018limits,afzal2020study,choi2021use}. As a result, the sim-to-real gap has hindered progress in robotics~\citep{zhao2020sim,salvato2021crossing,dulac2019challenges}.

With the development of generative models trained on large-scale video data~\citep{ho2022imagen,villegas2022phenaki,singer2022make}, recent work has shown that video world models can visually emulate interactions with the physical real world, by conditioning on control inputs in the form of text~\citep{yang2023learning,videoworldsimulators2024} or keyboard strokes~\citep{bruce2024genie}. This brings up an interesting question --- could video world models be used to emulate robot interactions with the real world, hence being used as an environment to evaluate robot policies in the world model before real-world testing or deployment?

Learning a dynamics model from past experience and performing rollouts in the learned dynamics model has been extensively studied in model-based reinforcement learning (RL)~\citep{hafner2019dream,fonteneau13batch,zhang2021autoregressive,kaiser2019model,yu2020mopo}. However, most of the existing work in model-based RL considers single-task settings, which puts itself at a disadvantage compared to model-free RL, since learning a dynamics model can be much harder than learning a policy in the single-task setting. Nevertheless, we make the important observation that
\begin{quoting}
    \textit{While there can be many tasks and policies, there is only \textbf{one physical world} in which we live that is governed by the \textbf{same set of physical laws}.} 
\end{quoting}
This makes it possible to learn a single world model that, in principle, can be used as an interactive environment to evaluate any policies on any tasks.

\begin{wrapfigure}{r}{6.4cm}
\centering
    \includegraphics[width=\linewidth]{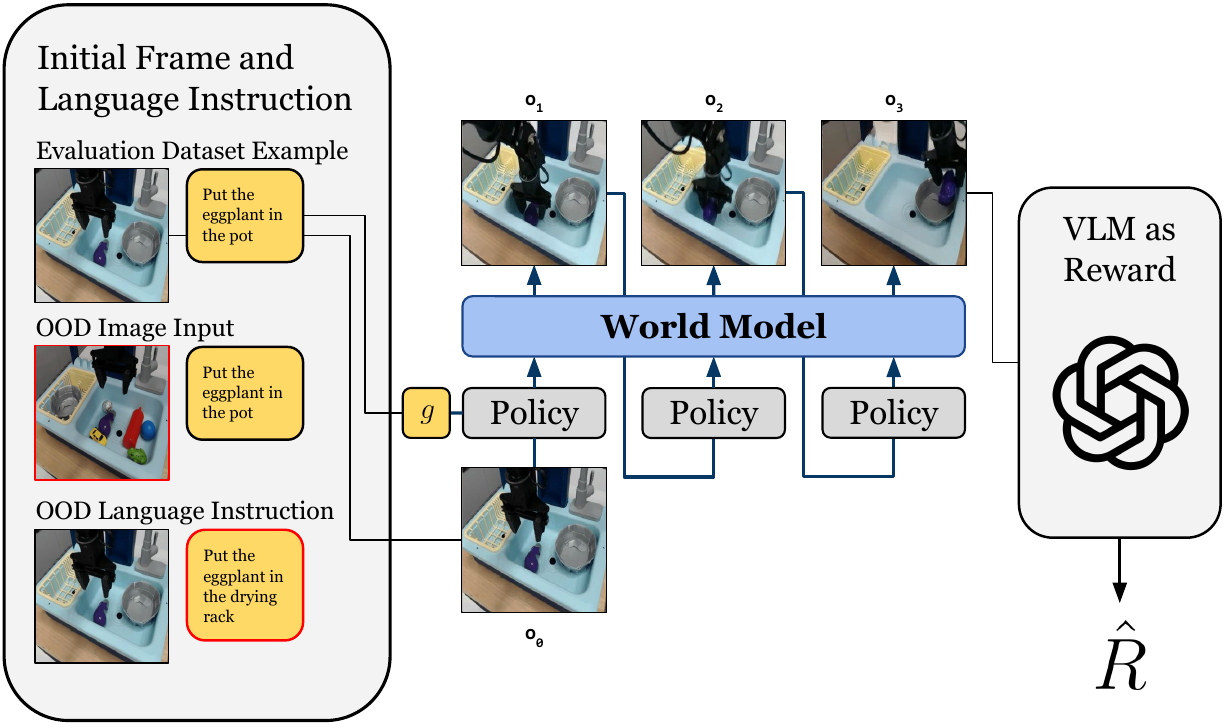}
\captionof{figure}{\textbf{Overview of \method.} Given an initial frame and an action sequence predicted by a policy, \method uses a world model to interactively predict future frames, serving as a generative simulator. \method then passes the generated rollout to a VLM which provides rewards. \method can easily be used to test policies on OOD tasks and environments by changing the input language instruction or directly modifying the initial image. 
}
\label{fig:eval-pipeline}
\end{wrapfigure}
Inspired by this observation, we propose a world-model-based policy evaluation environment (\method), as shown in Figure~\ref{fig:eval-pipeline}. \method first combines knowledge of the world across diverse environments by learning a \emph{single} world model that generates videos conditioned on actions.
To enable efficient rollouts of policies which predict different-length action chunks, \method aligns its diffusion horizon length with policies' chunk sizes at inference time.
With video rollouts from the world model, \method then uses a vision-language model (VLM) to determine tasks' success from generated videos.

Our experiments show that \method can emulate end-effector controls across different control axes highly effectively for robots with different morphologies. We then use the world model to evaluate VLA-based robot policies by rolling out the policies in the world model starting from real initial frames, and compare their success rates (policy values) in \method to those achieved in real-world experiments. Our result suggests that policy values in \method are highly correlated with policy performance in the real world, and the relative rankings of different policies are preserved.

Furthermore, as \method requires only a single initial frame as input, we show how we can easily design out-of-distribution (OOD) tasks and environments and then use \method to evaluate robot policies within these newly ``created'' environments. We find that modern robot policies still struggle to distinguish some classes of objects by their shape, and can even be distracted by adversarial facades of objects.

Although simulating realistic object interactions remains challenging, we believe \method can serve as a highly useful tool for sanity check and testing robot policies safely and reproducibly before deploying them on real robots.
Key contributions of this paper include:

\begin{itemize}[leftmargin=*, nosep]
\item We propose to use video world model to evaluate robot policies across different robot morphologies, and perform a comprehensive set of studies to understand its feasibility.
\item We propose flexibly aligning diffusion horizon length with policies' action chunk sizes for efficient rollouts of a variety of policies over hundreds of interactive steps.
\item We show a single world model learned on data from diverse tasks and environments can enable policy value estimates that highly correlate with real-world policy success rates.
\item We demonstrate the ease
of testing robot policies on OOD tasks and environments within an autoregressive video generation-based world model.
\end{itemize}

%% file: prelimiaries.tex
\section{Problem Formulation}
\label{sec:preliminaries}


In this section, we define relevant notations and review the formulation of offline policy evaluation (OPE). We also situate OPE in practical settings with partially observable environments and image-based observations.

\paragraph{Multi-Task POMDP.} We consider a multi-task, finite-horizon, partially observable Markov Decision Process (POMDP)~\citep{puterman2014markov,kaelbling1995partially}, specified by $\mathcal{M} = (S, A, O, G, R, T, \mathcal{E}, H)$, which consists of a state space, action space, observation space, goal space, reward function, transition function, emission function, and horizon length. A policy $\pi$ interacts with the environment for a goal starting from an initial state $g, s_0\sim G$, producing a distribution $\pi(\cdot|s_t, g)$ over $A$ from which an action $a_t$ is sampled and applied to the environment at each step $t\in [0, H]$. The environment produces a scalar reward $r_t = R(s_t, g)$, and transitions to a new state $s_{t+1}\sim T(s_t, a_t)$ and emits a new observation $o_{t+1}\sim\mathcal{E}(s_{t+1})$. We consider the sparse reward setting with $R(s_H, g)\in \{0, 1\}$ and $R(s_t, g) = 0, \forall t<H$, where $g$ is a language goal that defines the task. Data is logged from previous interactions into an offline dataset $D=\{g, s_0, o_0, a_0, ..., s_H, o_H, r_H\}$. The value of a policy $\pi$ can be defined as the total expected future reward:
\begin{align}
\rho(\pi) =& \mathbb{E}[R(s_H, g)|s_0, g\sim G, a_t\sim\pi(s_t, g),\nonumber\\  
&s_{t+1}\sim T(s_t, a_t), \forall t\in[0, H]].
\end{align}
Estimating the value of $\rho(\pi)$ from previously collected data $D$, known as offline policy evaluation (OPE)~\citep{levine2020offline}, has been extensively studied~\citep{thomas2016data,jiang2016doubly,fu2021benchmarks,yang2020off,thomas2015high}. However, existing work in OPE mostly focuses on simulated settings that are less practical (e.g., assumptions about full observability, access to ground truth states).

\paragraph{Model-Based Evaluation.} 

Motivated by characteristics of a real-robot system such as image based observations, high control frequencies, diverse offline data from different tasks/environments, and the lack of access to the ground truth state of the world, we consider the use of offline data to learn a \emph{single} world model $\hat T(\cdot|\mathbf{o}, \mathbf{a})$, where $\mathbf{o}$ represents a sequence of previous image observations and $\mathbf{a}$ represents a sequence of next actions. A sequence of next observations can be sampled from the world model $\mathbf{o'}\sim \hat T(\mathbf{o}, \mathbf{a})$.
With this world model, one can estimate the policy value $\rho(\pi)$ with Monte-Carlo sampling using stochastic rollouts from the policy and the world model:
\begin{align}
\hat\rho(\pi) =& \mathbb{E}[\hat R([o_0, ..., o_H], g)|s_0, g\sim G, \mathbf{a}\sim\pi(\mathbf{o}, g), \nonumber \\& \mathbf{o'}\sim \hat T(\mathbf{o}, \mathbf{a}), \mathbf{o} = \mathbf{o'}],
\end{align}
where $\hat R$ is a learned reward function. Previously, model-free policy evaluation may be more preferable since in a single task setting, dynamics models are potentially harder to learn than policy values themselves, and doing rollouts in a dynamics model may lead to compounding errors~\citep{xiao2019learning}. However, we make the key observations that while there can be many tasks and many policies, there is only one physical world that is governed by the same set of physical laws. 
As a result, learning a world model can benefit from diverse data from different tasks and environments with different state spaces, goals, and reward functions. More importantly, a world model can be directly trained on image-based observations, which is often the perception modality of real-world robots.

%% file: method.tex
\section{Building and Evaluating the World Model}
\label{sec:methods}

In this section, we first describe our implementation of world model training and inference. Then, we discuss how we validate our world model's performance prior to rolling out real robot policies within it in the next section.

\subsection{Building the World Model}

First, we describe the architecture and key implementation details, followed by our proposed inference scheme for policy rollouts.

\subsubsection{World Model Training}

We train a latent Diffusion Transformer \citep{peebles2023scalable} on sequences of frames paired with actions, using Diffusion Forcing \citep{chen2024diffusion} to enable autoregressive frame generation. Per-frame robot action vectors are linearly projected to the model dimension and added elementwise to diffusion timestep embeddings, the result of which is used to condition the model through AdaLN-Zero modulation, similar to class conditioning in \cite{peebles2023scalable}.
To ensure the world model is fully controllable by robot actions, we propose to randomly drop out actions for entire video clips, and use classifier-free guidance to improve the world model's adherence to action inputs. Conditioning on previous frames' latents is achieved via causal temporal attention blocks interleaved between spatial attention blocks, as in \cite{bruce2024genie, ma2025lattelatentdiffusiontransformer}. See Appendix \ref{app:architecture} for additional implementation details.


\begin{figure*}[t]
    \centering
    \includegraphics[width=\linewidth]{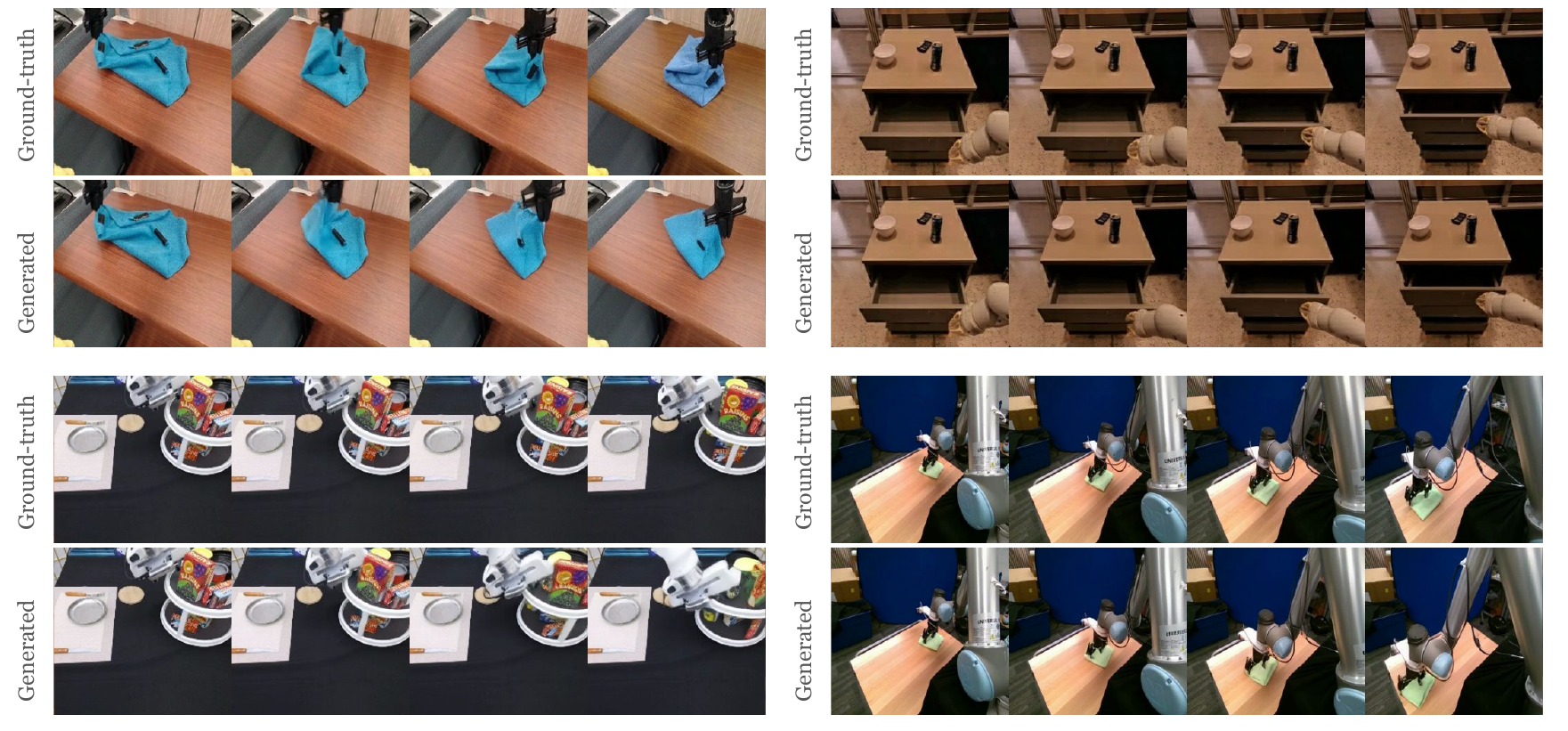}
    \caption{
    \textbf{Qualitative evaluation of the world model on Bridge, RT-1, VIOLA, and Berkeley UR5.} In each group, top row shows the ground truth video from the real robot. Bottom row shows the generated video from the world model conditioned on the \emph{same} actions as the original video. The world model closely follows the true dynamics across different robot morphologies.}
    \label{fig:real-qualitative}
\end{figure*}

\subsubsection{Rolling Out a Policy in the World Model}
\label{sec:rolling_out}

Our policy evaluation pipeline operates through an iterative loop between the robot policy and the world model. First, the world model is initialized with an initial observation $o_0$, which is then passed as input to a policy $\pi$ which produces a chunk of actions $\mathbf{a}_\text{pred}$. The actions are passed back to the world model, which predicts a new frame for each action in $\mathbf{a}_\text{pred}$. The latest frame produced by the world model is then returned to the policy as its next input observation.

Since different robot policies output a different number of actions at once~\citep{kim2406openvla,brohan2022rt,chi2023diffusion}, \method needs to support efficient prediction of a chunk of videos conditioned on a chunk of (variable length) actions. By virtue of being trained with Diffusion Forcing, as well as our usage of a causal temporal attention mask, we can flexibly control how many frames our world model denoises in parallel at inference time, i.e. its prediction \textit{horizon length}. We propose setting the horizon equal to the policy's action chunk size, $|\mathbf{a}_\text{pred}|$. This has the benefit of efficient frame generation for policies with differing action chunk sizes, all from a single world model checkpoint. This contrasts with prior diffusion world models for robotics, such as Cosmos \citep{nvidia2025cosmosworldfoundationmodel}, which, due to being trained with bidirectional attention and a fixed context length, must always denoise 16 latent frames in parallel. This constraint results in wasted compute for action chunk sizes less than the context length and unrealized parallelism for chunk sizes which are larger. On the other hand, our design allows parallelism to flexibly match the number of actions, thus utilizing hardware more effectively (see Appendix~\ref{app:parallelism}).

\subsubsection{VLM as Reward}

We opt for GPT-4o \citep{openai2024gpt4ocard} as a reward model, passing in the sequence of frames from the generated rollout and the language instruction (see the prompt for the VLM in Appendix \ref{app:vlm}). In certain cases where both policies being evaluated fail to perform a task end-to-end, it is still helpful to get signals on which policy is closer to completing a task. We can specify these partial credit criteria to the VLM to further distinguish performance between different policies, which has been done manually using heuristics in prior work~\citep{kim2406openvla}. We validate the accuracy of VLM-predicted rewards in Appendix~\ref{app:validate_vlm}.

\subsection{Evaluating the World Model}

Next, we describe how we validate the performance of our world model prior to policy evaluation, ensuring that it exhibits realistic robot movement and adheres to arbitrary action controls.

\begin{figure}[t]
    \centering
    \includegraphics[width=\linewidth]{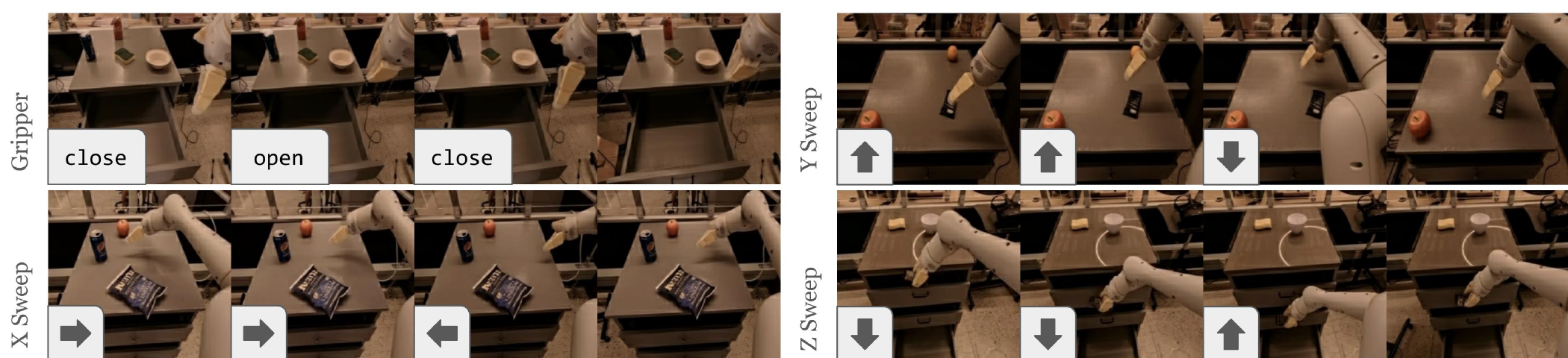}
    \vspace{-5mm}
    \caption{\textbf{Results on end-effector control} across action dimensions. Generated videos closely follow the gripper controls such as open and close the gripper as well as moving in different directions starting from any initial observation frame. Results for control sweeps on the Bridge robot can be found in Figure \ref{fig:qualitative-app-bridge} in Appendix~\ref{app:result-qualitative}.
    }
    \label{fig:gripper-control}
\end{figure}

\subsubsection{Agreement with Validation Split}

First, we test the world model's ability to generate similar videos as running a robot in the real world. Specifically, we take the validation split of initial images from the Open-X Embodiment dataset, and predict videos conditioned on the \emph{same} action sequences as in the original data. Figure~\ref{fig:real-qualitative} shows that the generated rollouts generally follow the real-robot rollouts across different initial observations and different robot morphologies.

\subsubsection{End-Effector Control Sweeps}

Next, we need a way to evaluate whether our world model can emulate arbitrary action sequences, beyond the kinds of action sequences present in the training data. We propose hard-coding a robot control policy by only moving one action dimension at once (and keeping the other action dimensions as zeros). The robot is then expected to move along that one action dimension with non-zero input, corresponding to moving in different horizontal and vertical directions as well as open and close its gripper. Figure~\ref{fig:gripper-control} shows that the generated videos faithfully follow the intended end-effector movement,\footnote{Results are best viewed as videos in the supplementary material.} despite the fact that these particular sequences of controls are not present in the training data.

%% file: experiment.tex
\section{Evaluating Policies in \method}
\label{sec:experiments}
Having established confidence in the world model's performance, 
we now use the world model to evaluate policies. We begin by rolling out three recent VLA policies in \method and check whether \method \textbf{reflects real-world success}.
(Section~\ref{sec:exp-corr}). We then assess whether \textbf{relative policy performance} is preserved, comparing different versions, sizes, and training stages of the same models (Section~\ref{sec:exp-relative}). Finally, 
we explore \method's potential to test policies on \textbf{out-of-distribution (OOD) tasks and environments} (Section~\ref{sec:exp-ood}), including novel instructions and altered visual contexts.

\subsection{Correlation between Real-World and Simulated Policy Performance} \label{sec:exp-corr}

\begin{figure}[t]
    \centering
    \begin{subfigure}{0.48\linewidth}
        \centering
        \includegraphics[width=\linewidth]{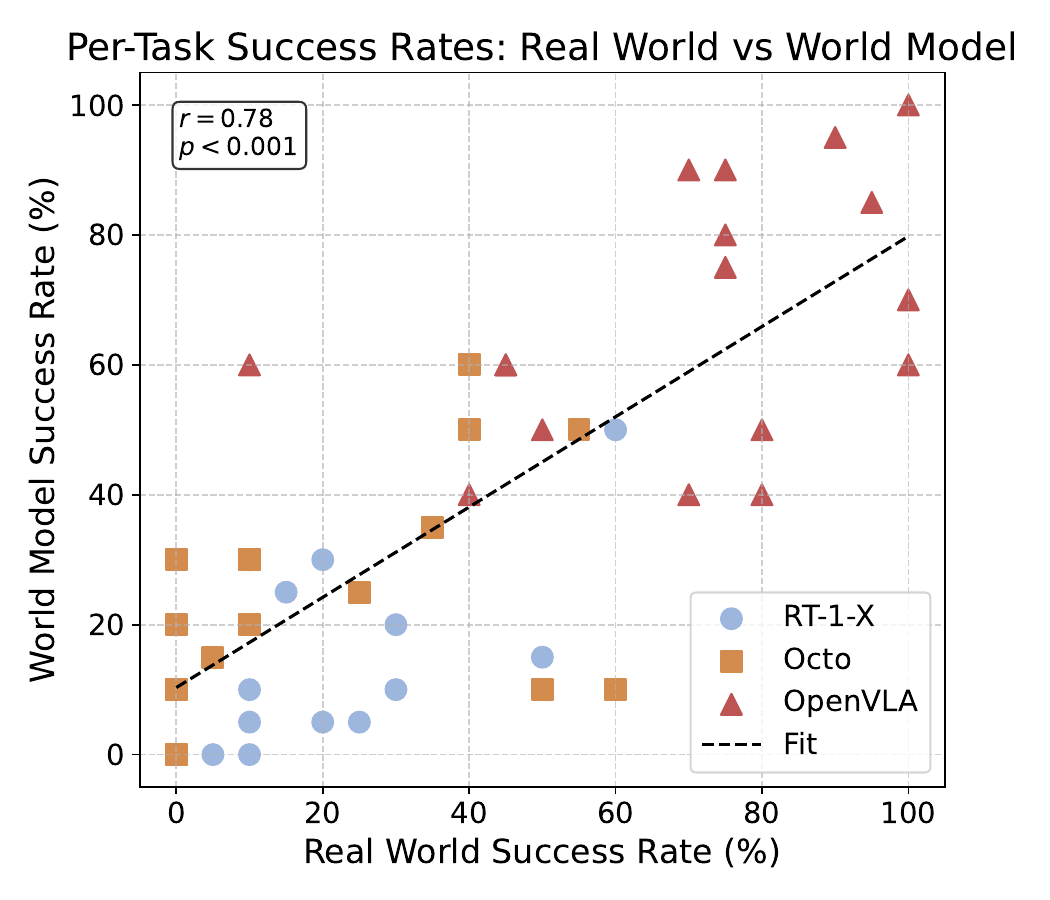}
        \caption{\textbf{Per-Task Task Success Rates.} Each point represents a task from Table \ref{app:tab:bridge_results_detailed}, with different policies being represented by different shaped markers. There is a strong correlation ($r=0.78$) between policy performance in our world model (y-axis) and within the real world (x-axis).}
        \label{fig:scatter_success_rates}
    \end{subfigure}
    \hfill
    \begin{subfigure}{0.48\linewidth}
        \centering
        \includegraphics[width=\linewidth]{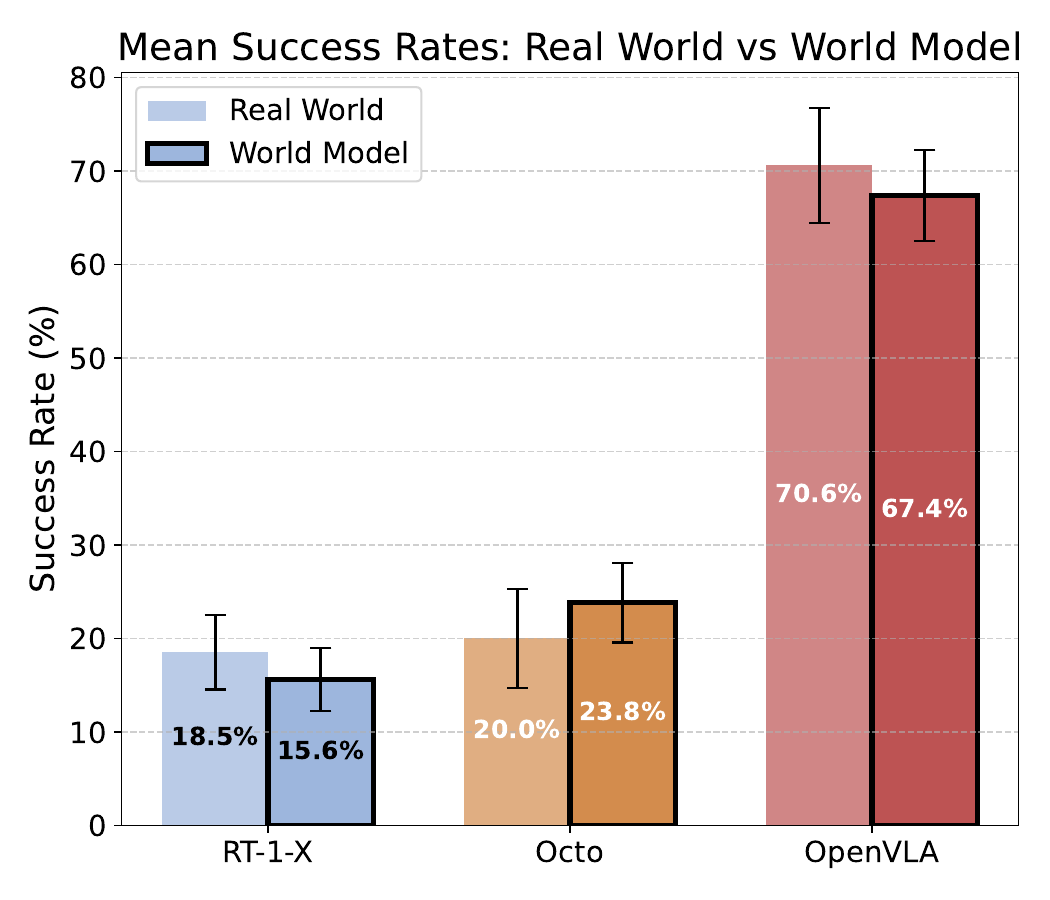}
        \caption{\textbf{Mean Success Rates.} Robot policies' mean success rates in the world model differ by an average of only 3.3\% between from the real world, near the standard error range for each policy. Relative performance rankings between RT-1-X, Octo, and OpenVLA are also preserved.}
        \label{fig:mean-success-rates-bar}
    \end{subfigure}
    \caption{Success rates of modern VLAs, as evaluated within \method{} and the real world.}
\end{figure}

\begin{figure}[t]
    \centering
    \includegraphics[width=\linewidth]{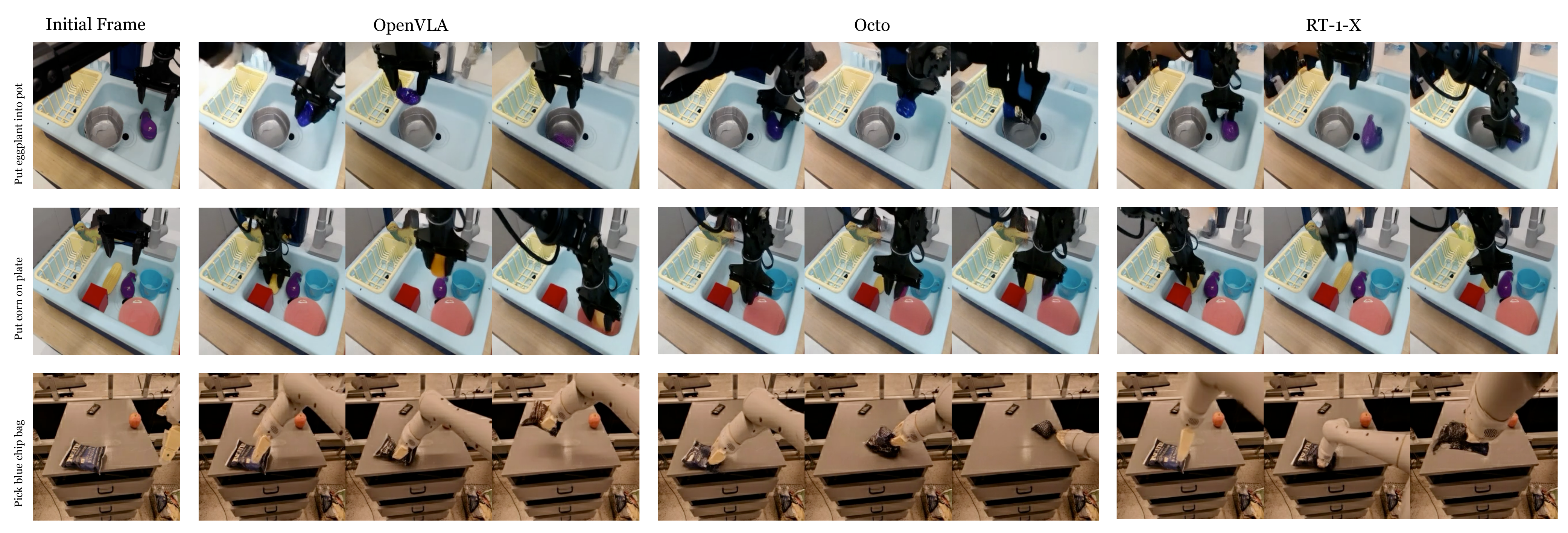}
    \caption{\textbf{Qualitative policy rollouts on Bridge and Google Robot} for RT-1-X, Octo, and OpenVLA. OpenVLA rollouts often lead to more visual successes than the other two policies across environments.}
    \label{fig:bridge_rollouts}    
\end{figure}

\paragraph{Qualitative Evaluation.} To ensure \method is useful for policy evaluation, we test whether policy performance within the world model is similar to that of the real world. To do so, we perform a direct comparison with the Bridge evaluation trials from OpenVLA \citep{kim2406openvla}. Specifically, the OpenVLA Bridge evaluation consists of 17 challenging tasks which are not present in the Bridge V2 \citep{walke2023bridgedata} dataset. We use \method to evaluate the three open-source policies evaluated in \cite{kim2406openvla}: RT-1-X \citep{o2023open}, Octo \citep{octo_2023}, and OpenVLA \citep{kim2406openvla}. For each task and each policy, \cite{kim2406openvla} perform 10 trials, each with randomized initial object locations. We obtain the first frame of the recorded rollouts for all trials of all tasks. We then simulate each of the 10 real-world trials by using the original initial frame to roll out the policy within the world model as described in Section $\ref{sec:rolling_out}$. We show qualitative rollouts in \method from different policies in Figure~\ref{fig:bridge_rollouts}, which shows that rollouts from OpenVLA generally perform better than rollouts from RT-1-X and Octo on the Bridge robot (top two rows). We further show that \method can be easily used to perform rollouts in other environments with other robots, such as the Google Robot (bottom row in Figure~\ref{fig:bridge_rollouts}).

\paragraph{Quantitative Evaluation.} Using the simulated rollouts from \method, we then compute the average task success rate similar to \cite{kim2406openvla}, and plot the success rate for each task for each policy in Figure~\ref{fig:scatter_success_rates}. We find that real-world task performance is strongly correlated with the task performance reported by the world model, achieving a Pearson correlation of $r = 0.78$. While per-task policy success rates within \method still differ slightly from those in the real world (see Table \ref{app:tab:bridge_results_detailed}), the mean success rates achieved by these policies within \method are quite close to the their real-world values, as shown in Figure~\ref{fig:mean-success-rates-bar}.
The success rates differ by an average of only 3.3\%, with RT-1-X achieving 18.5\% in the real world vs 15.5\% in the world model, Octo achieving 20.0\% vs 23.82\%, and OpenVLA achieving 70.6\% vs 67.4\%, respectively. See quantitative results of evaluating the three policies on the Google Robot in Appendix~\ref{app:google-robot}

\subsection{Policy Ranking within a World Model} \label{sec:exp-relative}

\begin{figure*}[t]
\centering
\begin{minipage}{0.48\textwidth}

    \centering
    \includegraphics[width=\linewidth]{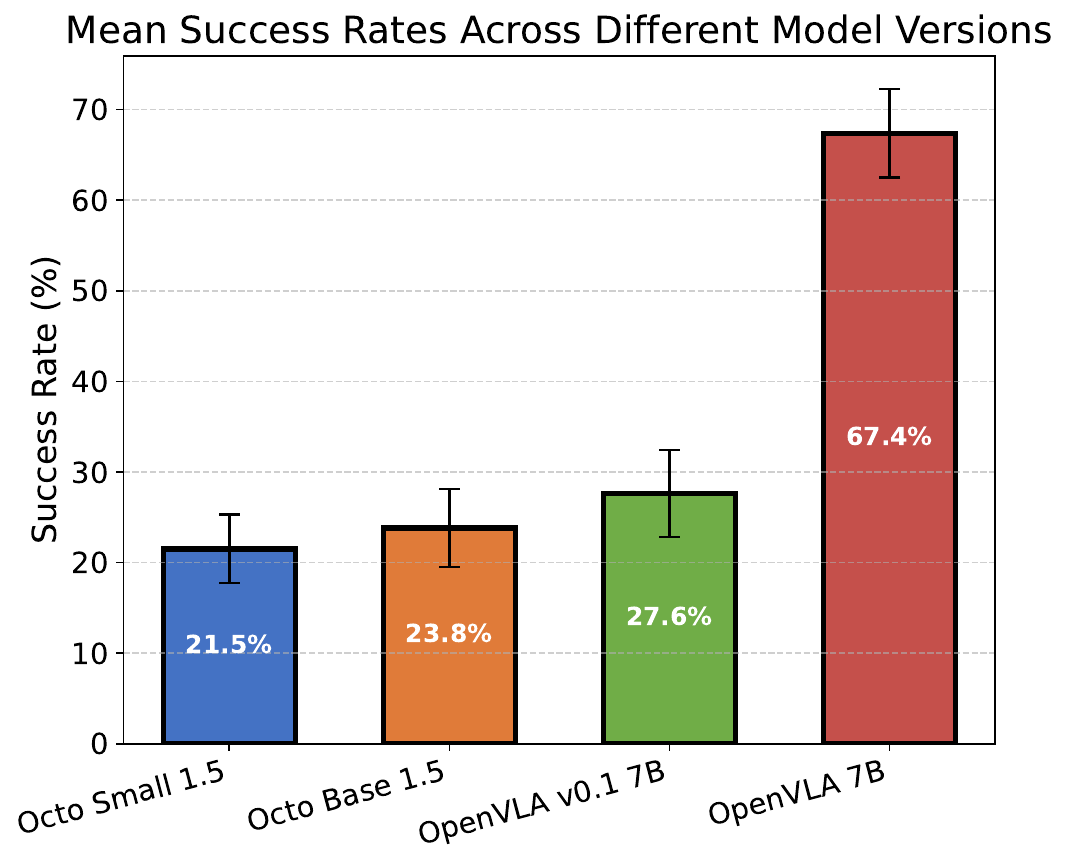}
    \caption{\textbf{Success Rates of different model versions in \method{}.} We evaluate different generations of Octo and OpenVLA in the world model, showing that \method{} assigns higher score to larger and more recent versions.}
    \label{fig:versions_vs_success_rates}
\end{minipage}\hfill
\begin{minipage}{0.48\textwidth}
    \centering
    \includegraphics[width=\linewidth]{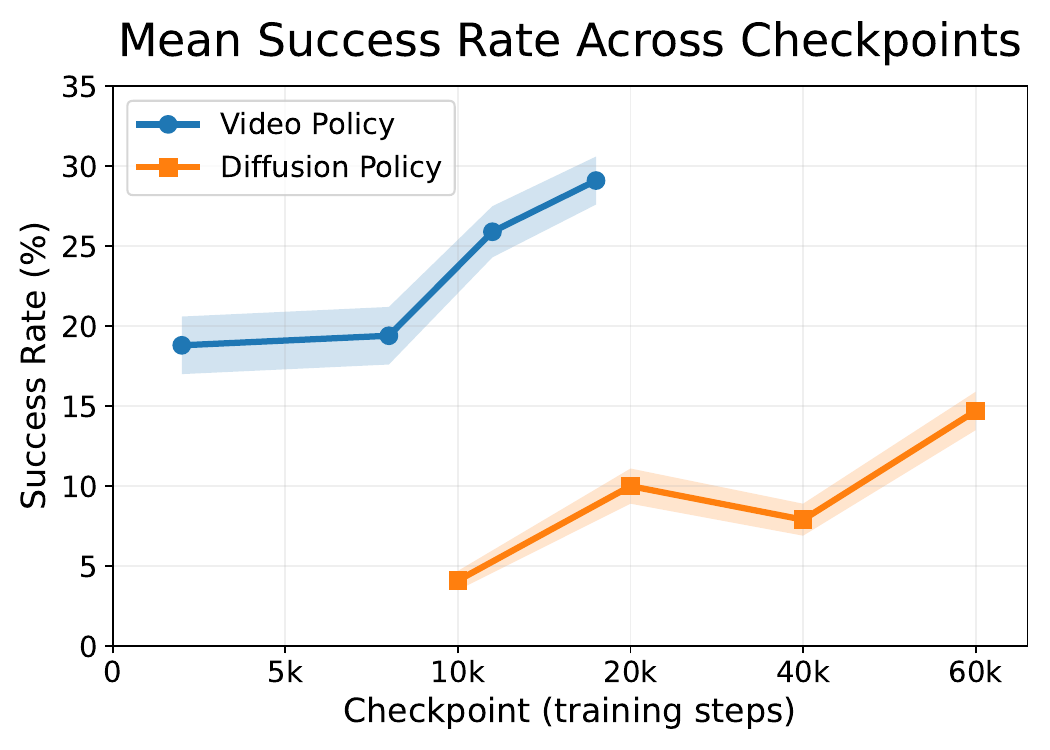}
    \caption{\textbf{Success Rate within \method{} throughout training.} We train a video-based policy and a diffusion policy from scratch and evaluate it within our world model as it trains. We see that mean task success rate within the world model increases with additional training steps.}
    \label{fig:steps_vs_success_rates}
\end{minipage}
\end{figure*}

Now we test whether \method can preserve policy rankings known a priori.
We evaluated policies
across different versions, sizes, and training stages
within \method{} on the OpenVLA Bridge evaluation task suite, and found their in-world-model performance rankings to be consistent with prior knowledge of their relative performance.

\textbf{Different VLAs with Known Ranking.} First, we average success rates across all 17 tasks and find that the relative performance rankings between RT-1-X, Octo, and OpenVLA are the same (Figure \ref{fig:mean-success-rates-bar}) within both \method and the real-world results reported in OpenVLA~\citep{kim2406openvla}.

\textbf{Same Policies across Versions and Sizes.}
We further examine whether \method{} preserves rankings between different versions and sizes of the same policy. In particular, we compare Octo-Small 1.5 against Octo-Base 1.5, and OpenVLA v0.1 7B, an undertrained development model, against OpenVLA 7B. As shown in Figure~\ref{fig:versions_vs_success_rates}, the larger and more recent models outperform their smaller or earlier counterparts within \method{}, consistent with the findings of real-world experiments performed in \cite{octo_2023} and \cite{kim2406openvla}. This provides additional evidence that \method{} faithfully maintains relative rankings even across model upgrades.

\textbf{Same Policy across Training Steps.} To examine whether \method{} provides meaningful signals for policy training, hyperparameter tuning, and checkpoint selections, we train two robot policies from scratch. Building on prior evidence of \method{}’s effectiveness in evaluating VLA-based policies, we extend our study to two additional families: a video prediction–based policy (UniPi)~\citep{du2023learninguniversalpoliciestextguided} and a diffusion-based policy (DexVLA)~\citep{wen2025dexvlavisionlanguagemodelplugin}, both trained on the Bridge V2 dataset (see Appendix~\ref{app:video_policy} and Appendix~\ref{app:diffusion_policy}). We evaluate checkpoints of the video prediction policy at 2K, 8K, 12K, and 18K steps, and the diffusion policy at 10K, 20K, 40K, and 60K steps.

As shown in Figure~\ref{fig:steps_vs_success_rates}, \method{} tends to assign higher success rates to checkpoints as they increase in training steps, consistent with the lower mean squared error these policies achieve on their validation splits. This demonstrates \method{}’s ability to preserve policy rankings across models with different amounts of training compute.

Thus, we have shown how \method{} can be used to obtain reasonable policy rankings. In particular, for the VLA-based policies we evaluate, we arrive at the same conclusions as real-world experiments about their relative performances. Notably, this is achieved all \textbf{without the manual effort} of setting up real robot evaluation environments and monitoring policy rollouts. While real-world evaluation can sometimes take days to complete, all \method{} rollouts reported here can be completed in under an hour on a single GPU and require only initial images for each trial. 

\subsection{Out-of-Distribution Inputs}\label{sec:exp-ood}

In this section, use \method{} to explore policies' performance on both OOD input images and OOD language instructions.

\begin{wrapfigure}{r}{6.4cm}
\centering
\includegraphics[width=\linewidth]{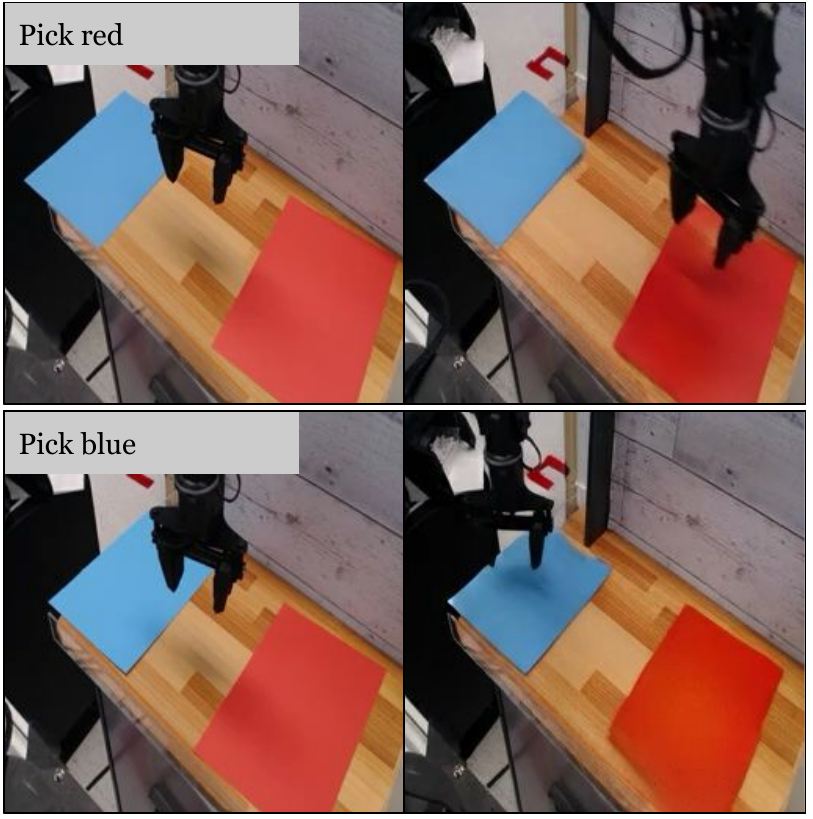}
\captionof{figure}{\textbf{OOD: Color Classification.} We add red and blue pieces of paper to a table, and ask the policies to ``pick red'' or ``pick blue'' (OOD image and language). 
OpenVLA excels, picking the correct colored paper in all trials, whereas all other policies score near chance.}
\label{fig:ood-image-classify-redblue}
\vspace{-5mm}
\end{wrapfigure}
\textbf{OOD Image Input.}
Using modern image generation models like Nano Banana \citep{SharonEtAl2025GeminiImageEditing}, we can easily generate new input images to initialize our world model with. We evaluate robot policies under three OOD settings: unseen object interaction, distractor objects, and object classification (see detailed results in Table~\ref{app:tab:ood_image_results_detailed}).

\begin{figure*}[t]
\centering
\begin{minipage}[t]{0.48\textwidth}
    \centering
    \includegraphics[width=\linewidth]{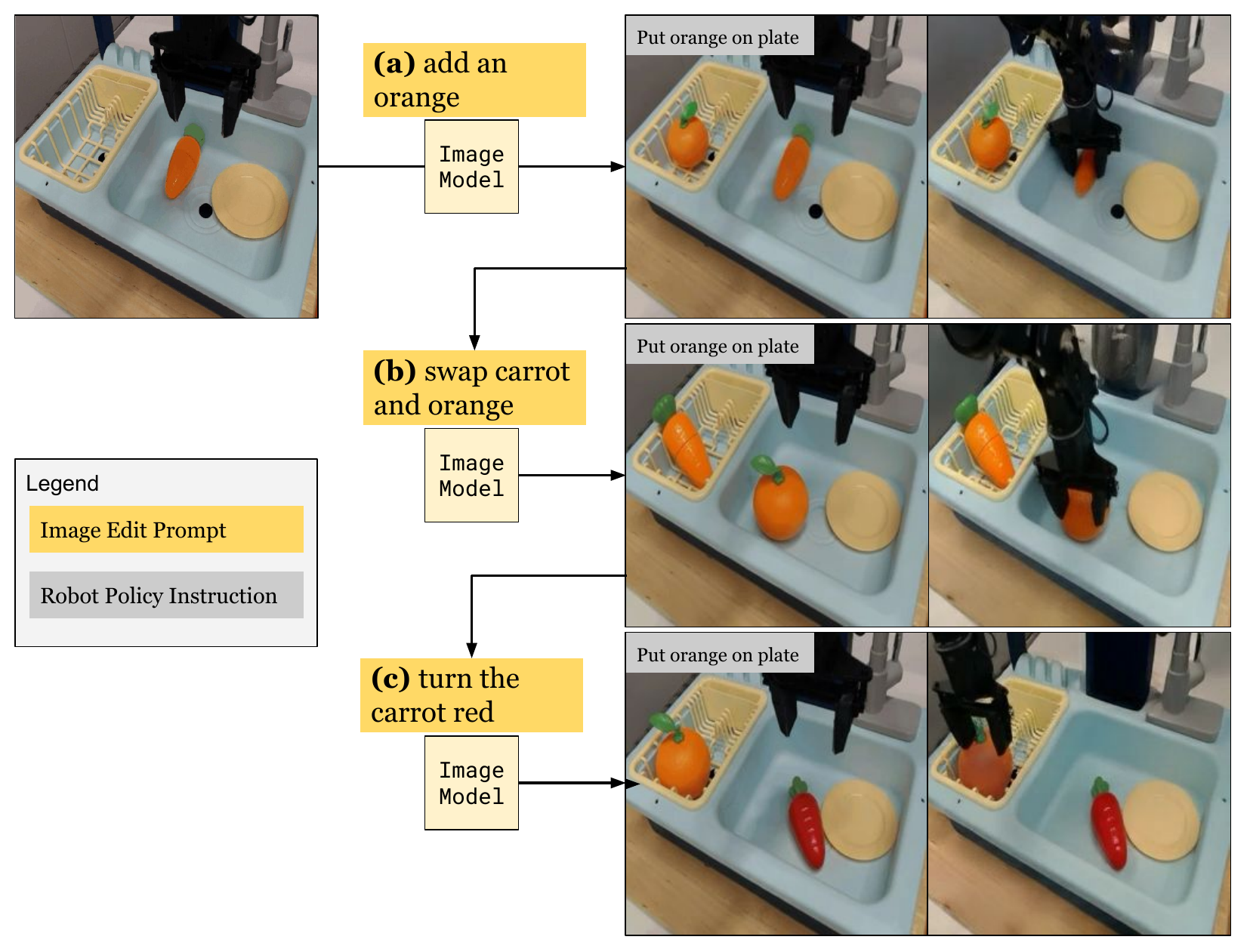}
    \caption{\textbf{OOD: Unseen object.} We use Nano Banana \citep{SharonEtAl2025GeminiImageEditing} to add an orange to the world model's initial frame. When both the orange and the carrot are present, (a-b) OpenVLA grabs whichever is closer. After (c) editing the carrot's color to red, however, the orange is correctly picked up.}
    \label{fig:ood_put_orange_on_plate}
\end{minipage}\hfill
\begin{minipage}[t]{0.48\textwidth}
    \centering
    \includegraphics[width=\linewidth]{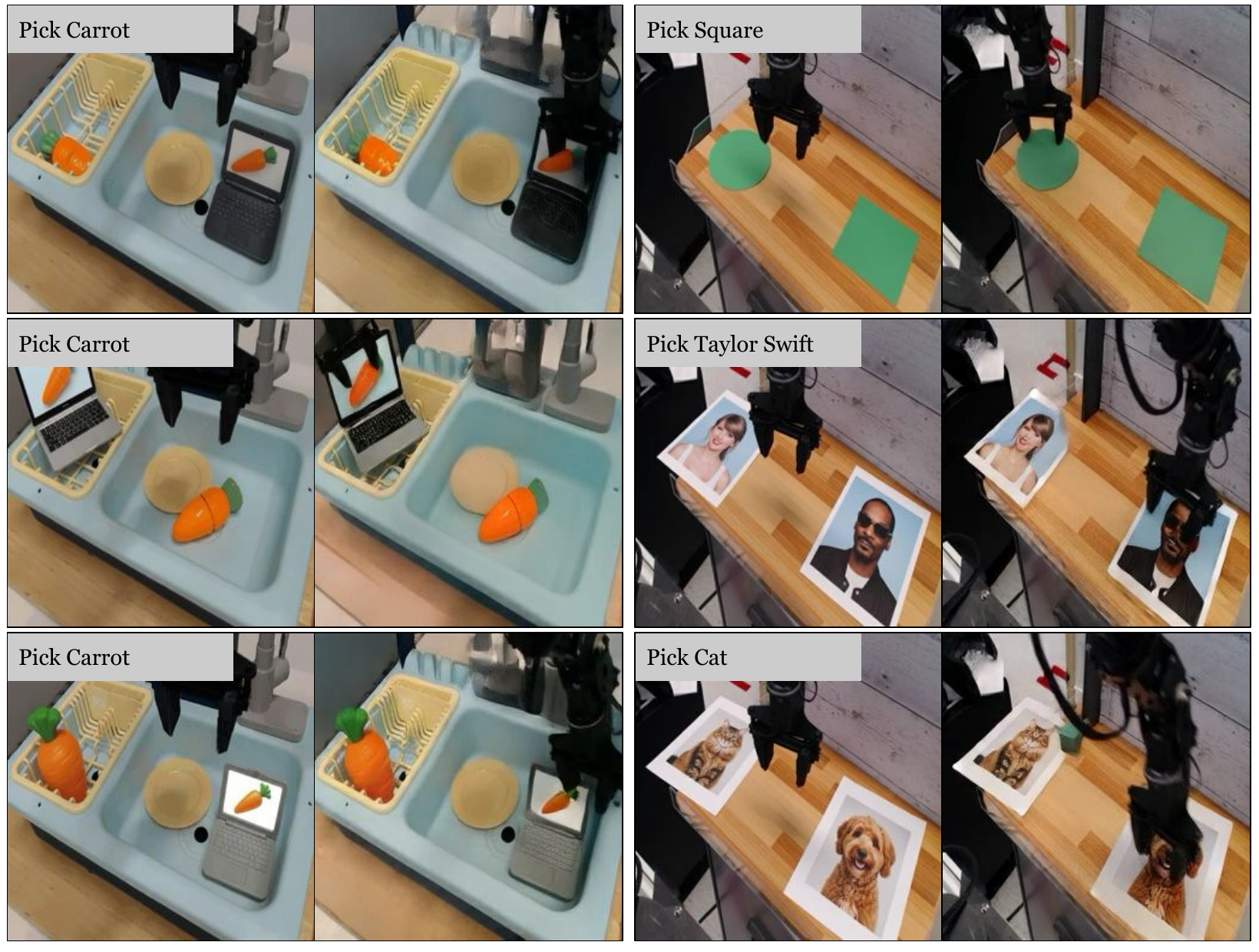}
    \caption{\textbf{OOD: Failure modes.} \textit{Left:} We add a laptop to the scene, which displays an image of a carrot. In 15\% of trials, OpenVLA grabs the laptop instead of the real carrot. \textit{Right:} We test the ability distinguish to between squares and circles, celebrity faces, and cats and dogs, with all policies scoring near-chance.}
    \label{fig:ood_advanced}
\end{minipage}
\end{figure*}

\begin{figure}[t]
\centering
\begin{minipage}{0.48\textwidth}
\centering
\scriptsize
\begin{tabular}{p{2.8cm} c c c}
\toprule
Task & RT-1-X & Octo & OpenVLA \\
\midrule
{\tiny Move Pot Into Drying Rack} & 3 & 0 & 7 \\
{\tiny Move The Pot To The Counter} & 0 & 0 & 1 \\
{\tiny Put Plate On Drying Rack} & 4 & 2 & 8 \\
{\tiny Put Yellow Corn In Red Cup} & 1 & 2 & 3 \\
\bottomrule
\end{tabular}
\captionof{table}{\textbf{Policy Evaluations Results on Bridge OOD Language Tasks.}
``Move the pot to the counter'' is perhaps the most challenging because the Bridge dataset does not contain trajectories which move objects outside of the sink basin. 
OpenVLA has the strongest performance, which we attribute to its more powerful language model backbone.}
\label{app:tab:bridge_ood_language}


\includegraphics[width=\linewidth]{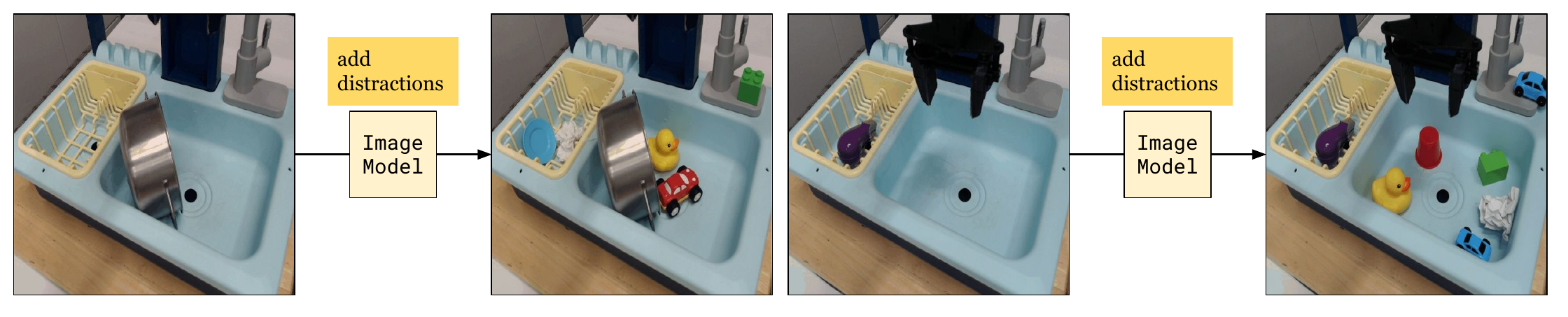}
\captionof{figure}{\textbf{OOD Distraction Examples.} We use Nano Banana \citep{SharonEtAl2025GeminiImageEditing} to add distractions to every image of the OpenVLA Bridge task suite. The resulting change in mean success rates can be seen in Figure~\ref{fig:world_model_vs_ood_success_rates}.}
\label{fig:ood_distraction_eval}
\end{minipage}\hfill
\begin{minipage}{0.48\textwidth}
\centering
\includegraphics[width=\linewidth]{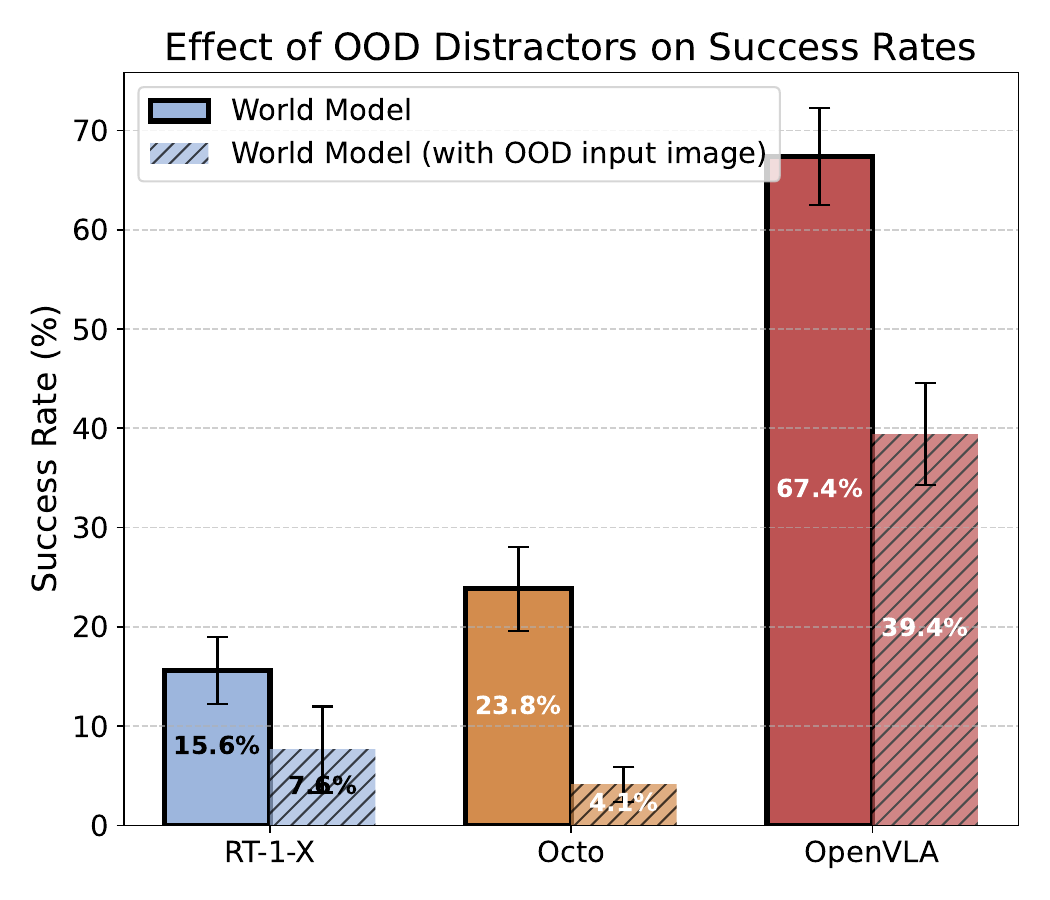}
\caption{\textbf{Effect of OOD Distractors.} We use an image editing model to add distractor objects to the Bridge evaluation suite, finding that RT-1-X drops in performance by 51\%, Octo by 83\%, and OpenVLA by 41.5\%, making OpenVLA the most robust to distractors. See Table~\ref{app:tab:distractor_results} for details.}
\label{fig:world_model_vs_ood_success_rates}
\end{minipage}
\end{figure}

\begin{itemize}[leftmargin=*,itemsep=0pt, after=\vspace{-5pt}]
    \item \textbf{Unseen Objects:} We edit a scene to contain both a carrot and an orange, asking the policy to pick up the orange (Figure \ref{fig:ood_put_orange_on_plate}). OpenVLA grabs whichever object is closer until we edit the carrot's color to be red, after which it always grabs the orange correctly. This suggests that it struggles to distinguish carrots and oranges by their shape.
    \item \textbf{Distractor Objects:} We use the image editing model to add a computer displaying an image of a carrot (Figure \ref{fig:ood_advanced}, left). We see that OpenVLA mistakenly to grabs the carrot on the computer screen in 15\% of trials, suggesting limited 3D/2D object distinction.
    \item \textbf{Classification:} We add a piece of paper on each side of a desk. We first color one paper red and the other blue and instruct the model to ``pick red''/``pick blue'' (Figure \ref{fig:ood-image-classify-redblue}). OpenVLA achieves a perfect score, always moving towards the correct color. Octo and RT-1, on the other hand, typically move towards whichever paper is closer, scoring no better than chance. 
    We also try more advanced classification tasks, (Figure~\ref{fig:ood_advanced}, right), but find that the policies all score near-chance.
\end{itemize}
For a more quantitative study, we modify all the initial frames of the OpenVLA's Bridge evaluation task suite to include random OOD distractor items (see Figure~\ref{fig:ood_distraction_eval}), keeping the language instructions the same. We then repeat the rollout procedure from Section~\ref{sec:exp-corr} in order to measure the degree to which the addition of unrelated objects affects policy performance. We find that all the tested VLAs degrade in performance, with OpenVLA being the most robust of the three (Figure~\ref{fig:world_model_vs_ood_success_rates}).

\begin{figure}[t]
    \centering
    \includegraphics[width=\linewidth]{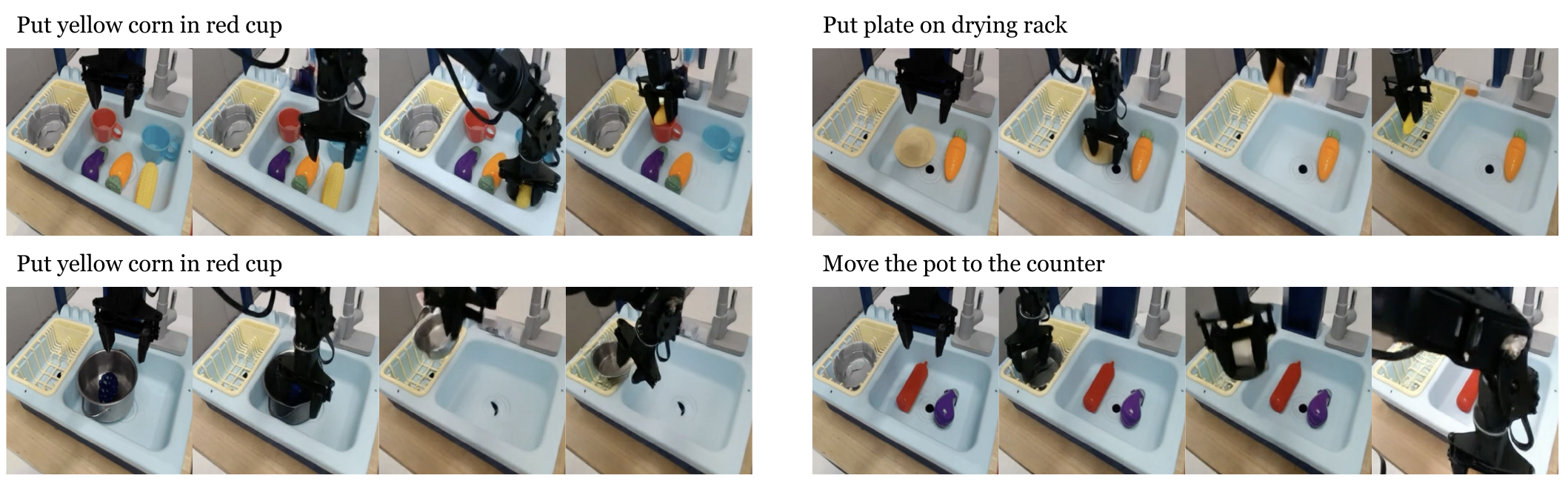}
    \caption{\textbf{OOD Language Instructions.} We pick a set of tasks from the OpenVLA Bridge evaluation suite and modify the language instruction, e.g. changing the the target object and/or its goal destination.}
    \label{fig:ood_language}    
\end{figure}

\textbf{OOD Language.} Additionally, even without access to an image editing model, we demonstrate that \method{} can be used to evaluate policies' performance on OOD language instructions. Starting from a set of initial frames from the tasks listed in Table \ref{app:tab:bridge_results_detailed}, we modify each task's language instruction, e.g. changing the target object and/or its goal state. Figure~\ref{fig:ood_language} shows rollouts from OpenVLA for these OOD language tasks.
We can then easily obtain success rates for these unseen tasks by rolling them out within \method{}, finding that OpenVLA generalizes best (see Table \ref{app:tab:bridge_ood_language}). Policies struggle across the board on the ``Move the pot to the counter'' task, with only OpenVLA achieving a single success. We suspect that OpenVLA consistently outperforms Octo and RT-1-X on OOD language tasks due to its strong VLM backbone and richer robot pretraining dataset \citep{kim2406openvla}.

The ability to use \method{} to quickly design and evaluate policies within OOD tasks and environments thus leads us to new findings about policies' strengths and weaknesses.
Future research could be prioritized to address these issues, all without spending extra effort to set up additional experiments in the real world or within handcrafted simulators.

%% file: related.tex
\section{Related Work}

\paragraph{Action-Conditioned Video Generation.} Previous work has shown that video generation can simulate real-world interactions~\citep{yang2023learning,videoworldsimulators2024}, robotic plans~\citep{du2024learning,du2023video}, and games~\citep{oasis2024,bruce2024genie,valevski2024diffusion,alonso2024diffusion} when conditioned on text or keyboard controls.
Prior work~\citep{nvidia2025cosmosworldfoundationmodel} has begun to  explore applying video generation to simulating complex robotic controls. We take this a step further by using video-based world models to quantitatively estimate robot policy success rates. \method draws architectural inspirations from prior work on video generation such as
Diffusion Forcing~\citep{chen2024diffusion} and Diffusion Transformers~\citep{peebles2023scalable}, but experiments with variable horizon lengths to support  efficient long-horizon rollouts for policies with a variety of action chunk sizes.

\paragraph{Policy Evaluation.} Off-policy and offline policy evaluation has long been studied in the RL literature~\citep{farajtabar2018more,jiang2015doubly,kallus2019double,munos2016safe,Precup00ET,Thomas15HCPE}. Some of these approaches are model-based, learning a dynamics model from previously collected data and rolling out the learned dynamics model for policy evaluation~\citep{fonteneau13batch,zhang2021autoregressive,yu2020mopo,hafner2020mastering}. Since learning a dynamics model is challenging and subject to accumulation of error, a broader set of work has focused on model-free policy evaluation, which works by estimating the value function~\citep{le2019batch,duan2020minimaxoptimal,sutton2009fast,sutton2016emphatic} or policy correction~\citep{kanamori2009least,nguyen2010estimating,nachum2019dualdice}.  \method performs model-based policy evaluation, but proposes to learn a single world model on image-based observation that can be used to evaluate different policies on different tasks. SIMPLER~\citep{li2024evaluating} aims to evaluate realistic policies by constructing software-based simulators from natural images and showed highly correlated curves between simulated evaluation and real-robot execution, but it is hard to evaluate OOD language and image input in SIMPLER without significant hand engineering of the software simulator. \citet{li2025worldeval} proposes to evaluate robot policies in a world model in a specific bi-manual manipulation setup, whereas \method focuses on evaluating policies across diverse environments and robot morphologies while enabling testing OOD language and image inputs.


%% file: conclusion.tex
\section{Conclusion}
\label{sec:conclusion}

We have presented \method, a world-model-based environment for evaluating robot policies. \method emulates realistic robot interactions and shows strong correlations between simulated evaluation and real-world policy outcomes. \method further provides the flexibility for evaluating OOD language instructions and performing tasks with an OOD initial frame. While not all interactions emulated by \method are fully realistic, \method serves as an important step towards safe and reproducible policy evaluation before deployment. 



%% file: appendix.tex
\begin{center}
{\huge Appendix}
\end{center}

\section{Additional Details of the Autoregressive Diffusion Transformer}\label{app:architecture}



\textbf{Implementation details:} We use the VAE from Stable Diffusion 3~\cite{esser2024scalingrectifiedflowtransformers} to independently encode 256$\times$256 image frames into latent space. We employ a 16-layer transformer with 1024 hidden dimensions and 16 attention heads. We train the world model on a diverse set of data sources, including 9 of the robot datasets from Open-X Embodiment whose action spaces can be unified, such as Bridge V2~\citep{walke2023bridgedata} and RT-1~\citep{brohan2022rt}. We encode actions from a 7-dimensional vector, using the 6-dimensional end-effector position and binary gripper state as our action space. Action spaces from different robots are aligned by normalizing each component's 10$^\text{th}$- and 90$^\text{th}$-percentile values to those of the RT-1 dataset. We train with a context length of 20 frames; for longer rollouts, we condition on a sliding window of the last 20 frames.

\begin{table}[ht]
\centering
\begin{tabular}{ll}
\toprule
\textbf{Hyperparameter} & \textbf{Value}  \\
\midrule
Total parameters & 609 M \\
Image Resolution & 256$\times$256 \\
DiT Patch Size & 2 \\
Input Channels & 16 \\
Hidden Size & 1024 \\
Layers & 16 \\
Attention Heads & 16 \\
MLP Ratio & 4 \\
Optimizer & AdamW (weight decay = $0.002$, $\beta_1 = 0.9, \beta_2 = 0.99$) \\
Learning rate & 8e-5 \\
Batch size & 16 \\
Action dimension & 7 \\
Training hardware & 2xA100 80GB \\
Training steps & 300k \\
Diffusion noise schedule & sigmoid \\
Sampling timesteps & 10 \\
Prediction target & $v$ \\
\bottomrule
\end{tabular}
\caption{Hyperparameters for training \method's video prediction model.}
\label{tab:hyper}
\end{table}

\begin{algorithm}
\caption{\method policy evaluation loop.}\label{alg:policy_eval_loop}
\begin{algorithmic}
\Require World model $\hat{T}$ with training context length $N_\text{train}$ and prediction horizon $h$, rollout length $N_\text{rollout}$, policy $\pi$ with action chunk size $|\mathbf{a}_\text{pred}|$, reward model $\hat{R}$, initial observation $o_0$, goal $g$
\State $\mathbf{o} \gets [o_0]$
\State $\mathbf{a} \gets [a_\text{null}]$
\State $n = 0$
\While{$n \leq N_\text{rollout}$}
\State $\mathbf{a_\text{pred}} \gets \pi(\mathbf{o}_n, g)$
\For{$i = 0$ to $\lceil |\mathbf{a}_\text{pred}| / h \rceil - 1$}

\State $\mathbf{a}_\text{ctx} \gets \mathbf{a}_{-N_\text{train}:}$
\State $\mathbf{o}_\text{ctx} \gets \mathbf{o}_{-N_\text{train}:}$
\State $\mathbf{o}_\text{pred} \gets \hat{T}(\mathbf{o}_\text{ctx}, \mathbf{a}_\text{ctx} || \mathbf{a}_{\text{pred}, h \cdot i : h \cdot (i+1)})$ \Comment{predict a block of $h$ frames in parallel}
\State $\mathbf{o} \gets \mathbf{o} || \mathbf{o}_\text{pred}$  \Comment{concatenate generated block of observation frames with observation history}
\State $\mathbf{a} \gets \mathbf{a} || \mathbf{a}_{\text{pred}, h \cdot i : h \cdot (i+1)}$
\EndFor
\State $n \gets n + n_\text{chunk}$
\EndWhile
\State $r \gets \hat{R}(\mathbf{o})$
\end{algorithmic}
\end{algorithm}

Algorithm~\ref{alg:policy_eval_loop} shows the detailed algorithm for performing a sliding window rollout of a policy with action chunk size $|\mathbf{a}_\text{pred}|$ and a world model with prediction horizon $h$. Note that in practice we always choose $h=|\mathbf{a}_\text{pred}|$ at inference time.

\clearpage
\newpage

\section{Details of VLM as Reward}\label{app:vlm}

\subsection{Prompt for VLM as Reward}
Prompt GPT-4o as Reward $\hat R$. Note that \texttt{has\_partial} is \texttt{True} if the chosen task has a partial credit criteria, which is the case for some tasks used in OpenVLA \citep{kim2406openvla}.

\begin{tcolorbox}
\begin{lstlisting}[breaklines=true]
Here is a sequence of frames from a robot policy which has been rolled out in a video-generation-based world model.
I need your help determining whether the policy is successful. How successfully does the robot complete the following task?

Instruction: {instruction}
{rubric.strip()}

Provide brief reasoning (2-3 sentences). Then output EXACTLY one final line:
Final Score: X
Where X is { 'one of 0, 0.5, or 1' if has_partial else '0 or 1' }.
No extra numbers after that line.
Note: Since this video was generated by a video prediction model (conditioned on robot actions), it may contain some artifacts due to the video model capacity.
\end{lstlisting}
\end{tcolorbox}

If there is a partial credit criteria, the rubric is:
\begin{tcolorbox}
\begin{lstlisting}[breaklines=true]
0   = Failure: little or no progress toward: "{instruction}"
0.5 = Partial: "{partial_desc}" achieved BUT the instruction not fully completed
1   = Success: Instruction fully completed (counts even if partial also true)
\end{lstlisting}
\end{tcolorbox}

Otherwise, the rubric is:

\begin{tcolorbox}
\begin{lstlisting}[breaklines=true]
Score rubric:
0 = Failure: instruction "{instruction}" not completed.
1 = Success: instruction completed.
\end{lstlisting}
\end{tcolorbox}

\subsection{Validating VLM Success Predictions}
\label{app:validate_vlm}

\begin{table}[b]
\centering
\footnotesize
\caption{\textbf{Performance of VLM as reward} (mean and standard error across 4 runs) on videos from RT-1~\citep{brohan2022rt} using ground truth task success labels. GPT-4o achieves high true positives and true negatives. Notably, GPT-4o as reward has very low false positive rate, which is especially important for not over-estimating a policy value.}
\label{tab:rt1-vlm-acc}
\begin{tabular}{l|c|c}
\toprule
    & RT-1 Success & RT-1 Fail \\
\midrule
VLM Success & 0.81 $\pm$ 0.14 (TP) & 0.03 $\pm$ 0.05 (FP) \\
VLM Fail & 0.19 $\pm$ 0.14 (FN) & 0.97 $\pm$ 0.05 (TN) \\
\bottomrule
\end{tabular}
\end{table}
To determine whether a VLM can serve as a reliable reward function, we pass rollout videos from the RT-1 dataset, along with the prompts constructed from the templates above, as inputs to query GPT-4o. We use whether the task is successful according to the RT-1 data (validation split) as the ground truth. Table~\ref{tab:rt1-vlm-acc} shows that GPT-4o achieves high true positive and true negative rate for real videos, indicating that it is an effective evaluator of task success. Notably, GPT-4o achieves very low false positives (i.e., the rollout is a failure but the VLM thinks it is a success), which is highly useful in policy evaluation.

\clearpage
\newpage
\clearpage
\newpage
\section{Architecture and Training Details of Video Based Policy}
\label{app:video_policy}

Our video-based policy follows the framework of UniPi~\cite{du2023learninguniversalpoliciestextguided}, combining a language-conditioned video prediction model with an inverse dynamics model. 

\begin{figure}[h]
    \centering
    \includegraphics[width=\linewidth]{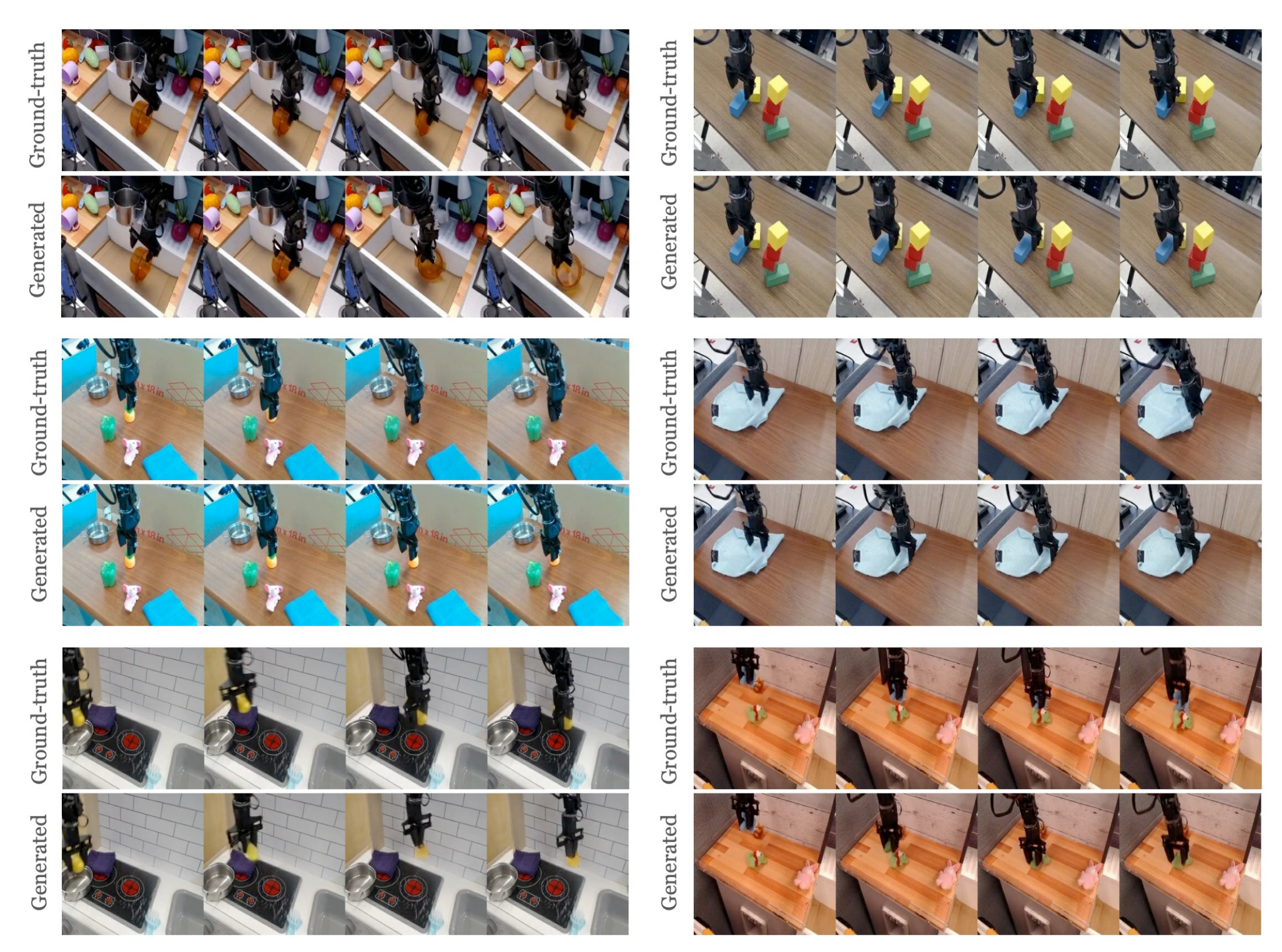}
    \caption{\textbf{Validation Visualization} of the language-conditioned video generation model on Bridge-V2. At inference, the model takes a UMT5-xxl instruction encoding and an initial frame, then predicts the next nine frames to complete the task.}
    \label{fig:video_prediction_viz}
\end{figure}

The video prediction module shares the same architecture as our world model, but replaces the conditioning on robot actions at each timestep with language instructions. For language conditioning, we employ the pretrained and frozen UMT5-xxl encoder~\cite{chung2023unimaxfairereffectivelanguage} to obtain token-level embeddings. These embeddings are aggregated via mean pooling to form a 4096-dimensional instruction representation. This representation is projected to match the model dimensionality and is used to modulate the diffusion transformer through adaptive layer normalization (adaLN-Zero). In this way, task semantics are directly integrated into the video prediction process. We train our video generation model for 180k steps on Bridge V2~\citep{walke2023bridgedata}. The visualization of the video generation policy on validation scenes can be seen in Figure~\ref{fig:video_prediction_viz}.

The inverse dynamics model predicts the action sequence given a short video clip of 10 frames. Each frame is encoded with a ResNet 50 backbone~\cite{he2015deepresiduallearningimage}, producing per-frame features $f_t$. To capture motion, we compute both $f_t$ and temporal differences $\Delta f_t = f_{t+1} - f_t$, concatenate them, and flatten across the clip. The resulting representation is passed through an MLP to predict $10\times d_a$ outputs, corresponding to the action dimension $d_a$ at each timestep. Input images are normalized with ImageNet statistics, and the model is trained with mean squared error on ground-truth actions. The inverse dynamics model is trained independently on the Bridge V2 dataset for 200k steps.

\subsection{Validation Visualization of Language Conditioned Video Generation Model}\label{app:video_prediction_viz}

\clearpage
\newpage
\section{Architecture and Training Details of Diffusion Policy}
\label{app:diffusion_policy}
We followed the recipe of DexVLA ~\citep{wen2025dexvlavisionlanguagemodelplugin} for training the diffusion policy. We load the Qwen2-VL-2B ~\citep{wang2024qwen2vlenhancingvisionlanguagemodels} backbone and the pre-trained control head, and perform an adaptation on BridgeV2 using LoRA \citep{hu2022lora}, which inserts low-rank adapter matrices inside the backbone’s attention and feed-forward blocks. These matrices along with the policy head are the only trainable modules in the adaptation stage. This preserves the backbone’s general vision-language competence, and makes adaptation compute and memory efficient. We fine-tune the model with adapters on Bridge-V2 for 60k steps. During training, we rescale the actions to $(-1, 1)$ to match the diffusion target range.

DexVLA’s policy head is trained as a denoising diffusion model with the standard $\epsilon$-prediction DDPM objective, i.e. at each update Gaussian noise is added to ground-truth action sequences at a randomly sampled diffusion step and the network is trained to predict that noise using an MSE loss. At inference, actions are generated with DDIM in a small number of steps, progressively denoising from a Gaussian initialization to a trajectory. We choose AdamW for the optimizer, using standard decoupled weight decay which applies decay to linear/attention weights, but exclude biases and LayerNorm parameters.

\clearpage
\newpage
\section{Additional Experimental Results}\label{app:exp}

\subsection{Additional Results on Real-Robot Videos}\label{app:result-qualitative}
\begin{figure}[h]
    \centering
    \includegraphics[width=\linewidth]{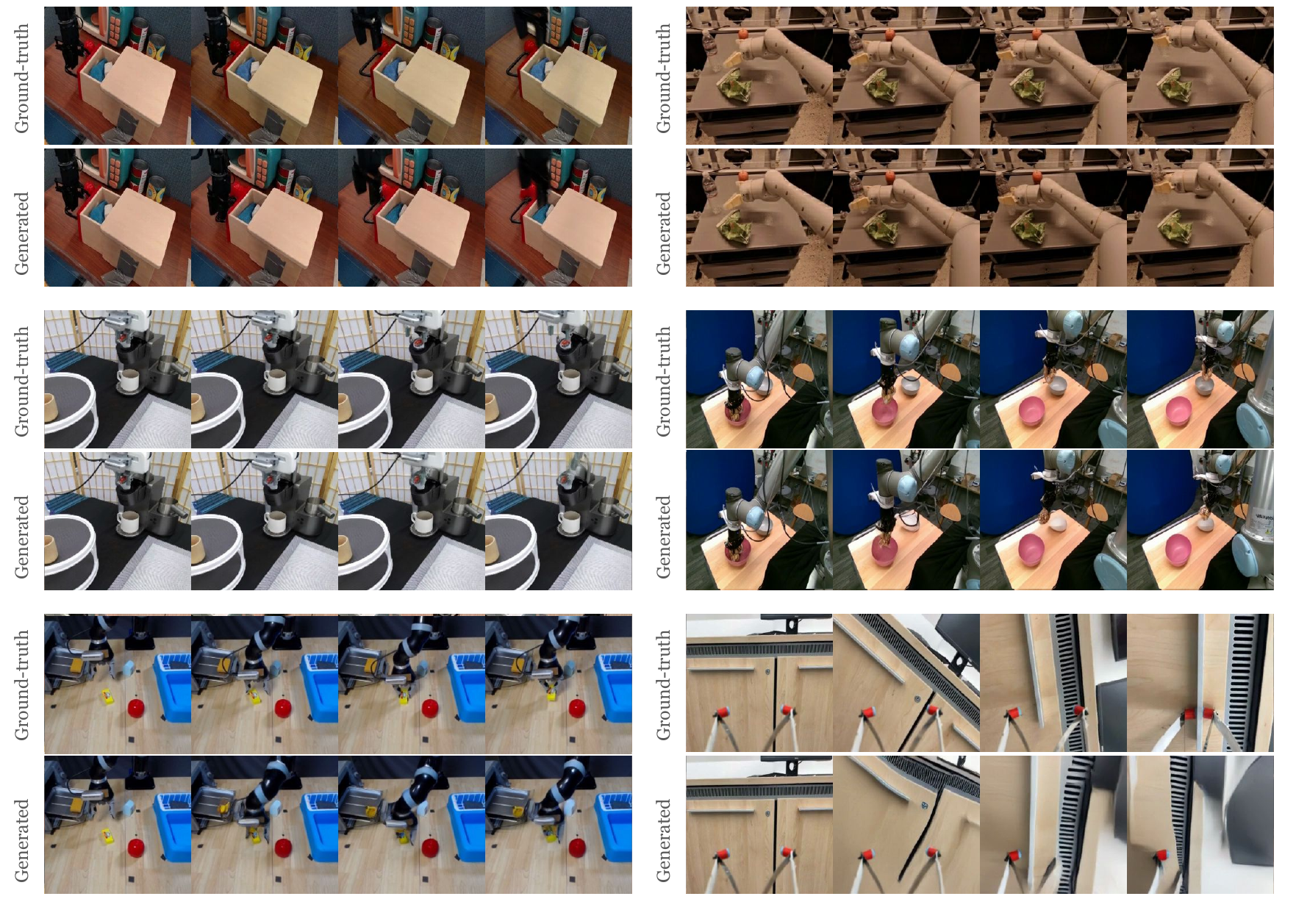}
    \caption{\textbf{Additional Qualitative Evaluation} of simulating actions from different robots. The world model generally generates the video that look very similar to the original video conditioned on the same actions that produced the original video in the real world.}
    \label{fig:qualitative-app}
\end{figure}

\begin{figure}[h]
    \centering
    \includegraphics[width=\linewidth]{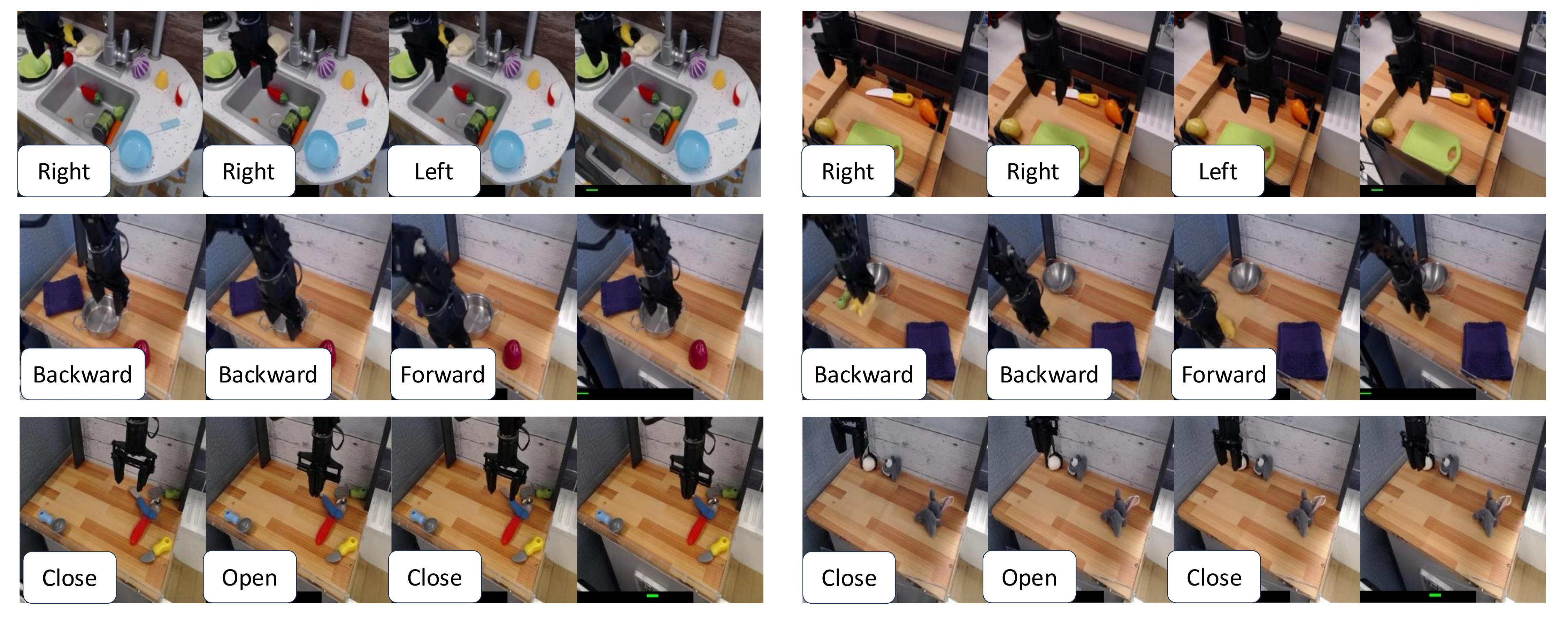}
    \caption{\textbf{Additional End-Effector Control Sweep on Bridge.} We simulate different gripper controls along different action dimensions corresponding to left-right, forward-backward, and gripper open-close. The world model generally generates videos that follow the actions.}
    \label{fig:qualitative-app-bridge}
\end{figure}



\clearpage
\newpage
\subsection{Additional Results on Google Robot}\label{app:google-robot}
To assess the generalizability of \method, we performed rollouts with different policies on Google Robot. For our analysis, we chose a subset of tasks from the RT-1 dataset \citep{brohan2022rt}. A partial score of 0.5 was assigned to a rollout if the robot attempted to reach the target location. OpenVLA again outperformed Octo and RT-1-X (see Table \ref{app:tab:fractal_results}). However, in this environment Octo and RT-1-X are narrowly behind. The strong performance of RT-1-X might be due to a higher proportion of Google Robot trajectories than WidowX in its pretraining mix.

\begin{table}[h]
\centering
\caption{\textbf{Policy rollouts on Google Robot} (RT-1 subset). OpenVLA outperforms RT-1-X and Octo, but by a smaller margin than on the Bridge dataset.}
\label{app:tab:fractal_results}
\resizebox{\textwidth}{!}{%
\begin{tabular}{l c c c c c}
\toprule
Task & \# Trials & \makecell{RT-1-X\\\# Successes} & \makecell{Octo\\\# Successes} & \makecell{OpenVLA\\\# Successes} \\
\midrule
Close Bottom Drawer & 10 & 9 & 8.5 & 6 \\
Open Left Fridge Door & 10 & 4.5 & 3.5 & 4 \\
Pick Blue Chip Bag & 10 & 5 & 5.5 & 9 \\
Place Redbull Can Upright & 10 & 1.5 & 3 & 3.5 \\
\bottomrule
\end{tabular}}
\end{table}


\clearpage
\newpage
\subsection{Detailed Results on the OpenVLA Bridge evaluation tasks}

\begin{table}[h]
\centering
\caption{\textbf{Detailed Bridge Evaluation Results} comparing RT-1-X \citep{o2023open}, Octo \citep{octo_2023}, and OpenVLA \citep{kim2406openvla} on the Bridge evaluation suite of tasks from \cite{kim2406openvla}. Real-world task success rates are taken directly from \citep{kim2406openvla}, \method{} success rates are from rolling out policies within our world model.}
\label{app:tab:bridge_results_detailed}
\resizebox{\textwidth}{!}{%
\begin{tabular}{l c cc cc cc}
\toprule
Task & \# Trials & \multicolumn{2}{c}{RT-1-X} & \multicolumn{2}{c}{Octo} & \multicolumn{2}{c}{OpenVLA} \\
\cmidrule(lr){3-4} \cmidrule(lr){5-6} \cmidrule(lr){7-8}
 &  & \makecell{Real-world\\\# Successes} & \makecell{\method{}\\\# Successes} & \makecell{Real-world\\\# Successes} & \makecell{\method{}\\\# Successes} & \makecell{Real-world\\\# Successes} & \makecell{\method{}\\\# Successes} \\
\midrule
Put Eggplant into Pot (Easy Version) & 10 & 1 & 1 & 5 & 1 & 10 & 7 \\
Put Eggplant into Pot & 10 & 0 & 0 & 1 & 2 & 10 & 6 \\
Put Cup from Counter into Sink & 10 & 1 & 3 & 1 & 3 & 7 & 9 \\
Put Eggplant into Pot (w/ Clutter) & 10 & 1 & 0.5 & 3.5 & 3.5 & 7.5 & 8 \\
Put Yellow Corn on Pink Plate & 10 & 1 & 3 & 4 & 6 & 9 & 9.5 \\
Lift Eggplant & 10 & 3 & 2 & 0.5 & 1.5 & 7.5 & 7.5 \\
Put Carrot on Plate (w/ Height Change) & 10 & 2 & 0.5 & 1 & 3 & 4.5 & 6 \\
Put Carrot on Plate & 10 & 1 & 0 & 0 & 1 & 8 & 4 \\
Flip Pot Upright & 10 & 2 & 3 & 6 & 1 & 8 & 5 \\
Lift AAA Battery & 10 & 0 & 1 & 0 & 0 & 7 & 4 \\
Move Skull into Drying Rack & 10 & 1 & 2 & 0 & 3 & 5 & 5 \\
Lift White Tape & 10 & 3 & 1 & 0 & 1 & 1 & 6 \\
Take Purple Grapes out of Pot & 10 & 6 & 5 & 0 & 2 & 4 & 4 \\
Stack Blue Cup on Pink Cup & 10 & 0.5 & 0 & 0 & 0 & 4.5 & 6 \\
Put \{Eggplant, Red Bottle\} into Pot & 10 & 2.5 & 0.5 & 4 & 5 & 7.5 & 9 \\
Lift \{Cheese, Red Chili Pepper\} & 10 & 1.5 & 2.5 & 2.5 & 2.5 & 10 & 10 \\
Put \{Blue Cup, Pink Cup\} on Plate & 10 & 5 & 1.5 & 5.5 & 5 & 9.5 & 8.5 \\
\midrule
Mean Success Rate & & 18.5$\pm$4.0\% & \cellcolor{lightlightgray}15.5$\pm$3.4\% & 20.0$\pm$5.3\% & \cellcolor{lightlightgray}23.82$\pm$4.3\% & 70.6$\pm$6.1\% & \cellcolor{lightlightgray}67.4$\pm$4.9\% \\
\bottomrule
\end{tabular}}
\end{table}

We report the mean success rate across tasks with standard error (SE) computed as
$$
\text{SE} = \frac{\mathrm{sd}(r_1,\dots,r_T)}{\sqrt{T}},
$$
where $r_i$ is the per-task success rate and $T$ is the number of tasks.

\clearpage
\newpage
\subsection{Detailed Results on OOD Image Evaluation Tasks}


\begin{table}[h]
\centering
\caption{\textbf{Detailed Bridge OOD Image task results.} OpenVLA appears to be more robust across the different OOD settings of object generalization, distractions and classification.}
\label{app:tab:ood_image_results_detailed}
\resizebox{\textwidth}{!}{%
\begin{tabular}{l l c c c c c}
\toprule
Category & Task & \# Trials & \makecell{RT-1-X\\\# Successes} & \makecell{Octo\\\# Successes} & \makecell{OpenVLA\\\# Successes} \\
\midrule
Object Generalization & Pick up Orange (Carrot closer to Gripper) & 10 & 1 & 1 & 4 \\
Object Generalization & Pick up Orange (Orange closer to Gripper) & 10 & 3 & 4 & 9 \\
Object Generalization & Pick up Orange (Replace Carrot with Radish) & 10 & 1 & 4 & 10 \\
Distractor Robustness & Pick up Carrot (With Computer on side) & 10 & 6 & 7 & 9 \\
Distractor Robustness & Pick up Carrot (Computer closer to gripper) & 10 & 3 & 1 & 8 \\
Classification & Pick \{Red, Blue\} & 20 & 8 & 10 & 20 \\
Classification & Pick \{Circle, Square\} & 20 & 8 & 10 & 12 \\
Classification & Pick \{Taylor Swift, Snoop Dogg\} & 20 & 7 & 10 & 11 \\
\bottomrule
\end{tabular}}
\end{table}

\begin{table}[h]
\centering
\caption{\textbf{Policy rollout performance comparison in the presence of unrelated distractions.} OpenVLA is more robust to distractions over RT-1-X and Octo. However, all policies suffer significant performance drop in the presence of distractors.}
\label{app:tab:distractor_results}
\resizebox{\textwidth}{!}{%
\begin{tabular}{l c c c c}
\toprule
Task & \# Trials & \makecell{RT-1-X\\\# Successes} & \makecell{Octo\\\# Successes} & \makecell{OpenVLA\\\# Successes} \\
\midrule
Put Eggplant into Pot (Easy Version) & 10 & 1 & 1 & 3 \\
Put Eggplant into Pot & 10 & 0 & 0 & 6 \\
Put Cup from Counter into Sink & 10 & 0 & 1 & 8 \\
Put Eggplant into Pot (w/ Clutter) & 10 & 0 & 0 & 4 \\
Put Yellow Corn on Pink Plate & 10 & 0 & 0 & 2 \\
Lift Eggplant & 10 & 0 & 1 & 7 \\
Put Carrot on Plate (w/ Height Change) & 10 & 0 & 0 & 2 \\
Put Carrot on Plate & 10 & 0 & 2 & 4 \\
Flip Pot Upright & 10 & 0 & 0 & 0 \\
Lift AAA Battery & 10 & 1 & 0 & 1 \\
Move Skull into Drying Rack & 10 & 3 & 0 & 5 \\
Lift White Tape & 10 & 0 & 0 & 4 \\
Take Purple Grapes out of Pot & 10 & 7 & 0 & 3 \\
Stack Blue Cup on Pink Cup & 10 & 0 & 0 & 4 \\
Put \{Eggplant, Red Bottle\} into Pot & 10 & 0 & 0 & 5 \\
Lift \{Cheese, Red Chili Pepper\} & 10 & 1 & 2 & 6 \\
Put \{Blue Cup, Pink Cup\} on Plate & 10 & 0 & 0 & 3 \\
\midrule
Mean Success Rate & & 7.6$\pm$4.3\% & 4.1$\pm$1.7\% & 39.4$\pm$5.1\% \\
\bottomrule
\end{tabular}}
\end{table}

\clearpage
\newpage
\section{Ablation Studies}\label{app:ablations}

\subsection{Dataset Size Analysis}

\begin{table}[h]
\centering
\footnotesize
\setlength{\tabcolsep}{2pt}
\caption{\textbf{Dataset ablation.} Larger training dataset improves all three metrics comparing generated videos and ground-truth validation videos. $\uparrow$ means higher the better.}
\label{tab:ablation-data}
\begin{tabular}{l|c|c}
\toprule
    & Subset (Bridge V1) & Full (Bridge V2)  \\
\midrule
 MSE $\downarrow$ &  0.015 & 0.010  \\
 LPIPS $\downarrow$ & 0.131 & 0.073  \\
 SSIM $\uparrow$ & 0.735 & 0.827 \\
\bottomrule
\end{tabular}
\end{table}

We measure MSE, LPIPS, and SSIM on generated videos from a model that is trained on less video data (Bridge V1~\citep{ebert2021bridge}) and compare with a model that is trained on more data (Bridge V2~\citep{walke2023bridgedata}). Table~\ref{tab:ablation-data} shows that the model trained on more data leads to improvements in all three metrics.

\subsection{Parallelism Efficiency Analysis}\label{app:parallelism}

\begin{table}[h]
\centering
\caption{\textbf{Parallelism efficiency comparison.} Inference time for generating 40-frame video rollouts on an A100 GPU with different horizon lengths, demonstrating the efficiency gains from parallel frame denoising.}
\label{tab:parallelism-efficiency}
\begin{tabular}{l|c}
\toprule
Prediction Horizon & Time (s) \\
\midrule
$h = 1$ & 93 \\
$h = 4$ & 33 \\
\bottomrule
\end{tabular}
\end{table}

Increasing the horizon from 1 to 4 frames achieves a 2.8× speedup. This is particularly useful for evaluating robot policies with differing action chunk sizes. For instance, OpenVLA~\citep{kim2406openvla} predicts just a single action per frame, while Octo~\citep{octo_2023} predicts 4 actions per frame. Using the same world model checkpoint, we can improve the efficiency of rollout generations by matching the horizon to the policy's chunk size at inference time.

%% file: iclr2026_conference.bbl
\begin{thebibliography}{73}
\providecommand{\natexlab}[1]{#1}
\providecommand{\url}[1]{\texttt{#1}}
\expandafter\ifx\csname urlstyle\endcsname\relax
  \providecommand{\doi}[1]{doi: #1}\else
  \providecommand{\doi}{doi: \begingroup \urlstyle{rm}\Url}\fi

\bibitem[Afzal et~al.(2020)Afzal, Katz, Goues, and Timperley]{afzal2020study}
Afsoon Afzal, Deborah~S Katz, Claire~Le Goues, and Christopher~S Timperley.
\newblock A study on the challenges of using robotics simulators for testing.
\newblock \emph{arXiv preprint arXiv:2004.07368}, 2020.

\bibitem[AI et~al.(2024)AI, Quevedo, McIntyre, Campbell, Chen, and Wachen]{oasis2024}
Decart AI, Julian Quevedo, Quinn McIntyre, Spruce Campbell, Xinlei Chen, and Robert Wachen.
\newblock Oasis: A universe in a transformer.
\newblock 2024.
\newblock URL \url{https://oasis-model.github.io/}.

\bibitem[Alonso et~al.(2024)Alonso, Jelley, Micheli, Kanervisto, Storkey, Pearce, and Fleuret]{alonso2024diffusion}
Eloi Alonso, Adam Jelley, Vincent Micheli, Anssi Kanervisto, Amos Storkey, Tim Pearce, and Fran{\c{c}}ois Fleuret.
\newblock Diffusion for world modeling: Visual details matter in atari.
\newblock \emph{arXiv preprint arXiv:2405.12399}, 2024.

\bibitem[Brohan et~al.(2022)Brohan, Brown, Carbajal, Chebotar, Dabis, Finn, Gopalakrishnan, Hausman, Herzog, Hsu, et~al.]{brohan2022rt}
Anthony Brohan, Noah Brown, Justice Carbajal, Yevgen Chebotar, Joseph Dabis, Chelsea Finn, Keerthana Gopalakrishnan, Karol Hausman, Alex Herzog, Jasmine Hsu, et~al.
\newblock Rt-1: Robotics transformer for real-world control at scale.
\newblock \emph{arXiv preprint arXiv:2212.06817}, 2022.

\bibitem[Brooks et~al.(2024)Brooks, Peebles, Holmes, DePue, Guo, Jing, Schnurr, Taylor, Luhman, Luhman, et~al.]{videoworldsimulators2024}
Tim Brooks, Bill Peebles, Connor Holmes, Will DePue, Yufei Guo, Li~Jing, David Schnurr, Joe Taylor, Troy Luhman, Eric Luhman, et~al.
\newblock Video generation models as world simulators. 2024.
\newblock \emph{URL https://openai. com/research/video-generation-models-as-world-simulators}, 3, 2024.

\bibitem[Bruce et~al.(2024)Bruce, Dennis, Edwards, Parker-Holder, Shi, Hughes, Lai, Mavalankar, Steigerwald, Apps, et~al.]{bruce2024genie}
Jake Bruce, Michael~D Dennis, Ashley Edwards, Jack Parker-Holder, Yuge Shi, Edward Hughes, Matthew Lai, Aditi Mavalankar, Richie Steigerwald, Chris Apps, et~al.
\newblock Genie: Generative interactive environments.
\newblock In \emph{Forty-first International Conference on Machine Learning}, 2024.

\bibitem[Chen et~al.(2024)Chen, Monso, Du, Simchowitz, Tedrake, and Sitzmann]{chen2024diffusion}
Boyuan Chen, Diego~Marti Monso, Yilun Du, Max Simchowitz, Russ Tedrake, and Vincent Sitzmann.
\newblock Diffusion forcing: Next-token prediction meets full-sequence diffusion.
\newblock \emph{arXiv preprint arXiv:2407.01392}, 2024.

\bibitem[Chi et~al.(2023)Chi, Xu, Feng, Cousineau, Du, Burchfiel, Tedrake, and Song]{chi2023diffusion}
Cheng Chi, Zhenjia Xu, Siyuan Feng, Eric Cousineau, Yilun Du, Benjamin Burchfiel, Russ Tedrake, and Shuran Song.
\newblock Diffusion policy: Visuomotor policy learning via action diffusion.
\newblock \emph{The International Journal of Robotics Research}, pp.\  02783649241273668, 2023.

\bibitem[Choi et~al.(2021)Choi, Crump, Duriez, Elmquist, Hager, Han, Hearl, Hodgins, Jain, Leve, et~al.]{choi2021use}
HeeSun Choi, Cindy Crump, Christian Duriez, Asher Elmquist, Gregory Hager, David Han, Frank Hearl, Jessica Hodgins, Abhinandan Jain, Frederick Leve, et~al.
\newblock On the use of simulation in robotics: Opportunities, challenges, and suggestions for moving forward.
\newblock \emph{Proceedings of the National Academy of Sciences}, 118\penalty0 (1):\penalty0 e1907856118, 2021.

\bibitem[Chung et~al.(2023)Chung, Constant, Garcia, Roberts, Tay, Narang, and Firat]{chung2023unimaxfairereffectivelanguage}
Hyung~Won Chung, Noah Constant, Xavier Garcia, Adam Roberts, Yi~Tay, Sharan Narang, and Orhan Firat.
\newblock Unimax: Fairer and more effective language sampling for large-scale multilingual pretraining, 2023.
\newblock URL \url{https://arxiv.org/abs/2304.09151}.

\bibitem[Du et~al.(2023{\natexlab{a}})Du, Yang, Dai, Dai, Nachum, Tenenbaum, Schuurmans, and Abbeel]{du2023learninguniversalpoliciestextguided}
Yilun Du, Mengjiao Yang, Bo~Dai, Hanjun Dai, Ofir Nachum, Joshua~B. Tenenbaum, Dale Schuurmans, and Pieter Abbeel.
\newblock Learning universal policies via text-guided video generation, 2023{\natexlab{a}}.
\newblock URL \url{https://arxiv.org/abs/2302.00111}.

\bibitem[Du et~al.(2023{\natexlab{b}})Du, Yang, Florence, Xia, Wahid, Ichter, Sermanet, Yu, Abbeel, Tenenbaum, et~al.]{du2023video}
Yilun Du, Mengjiao Yang, Pete Florence, Fei Xia, Ayzaan Wahid, Brian Ichter, Pierre Sermanet, Tianhe Yu, Pieter Abbeel, Joshua~B Tenenbaum, et~al.
\newblock Video language planning.
\newblock \emph{arXiv preprint arXiv:2310.10625}, 2023{\natexlab{b}}.

\bibitem[Du et~al.(2024)Du, Yang, Dai, Dai, Nachum, Tenenbaum, Schuurmans, and Abbeel]{du2024learning}
Yilun Du, Sherry Yang, Bo~Dai, Hanjun Dai, Ofir Nachum, Josh Tenenbaum, Dale Schuurmans, and Pieter Abbeel.
\newblock Learning universal policies via text-guided video generation.
\newblock \emph{Advances in Neural Information Processing Systems}, 36, 2024.

\bibitem[Duan \& Wang(2020)Duan and Wang]{duan2020minimaxoptimal}
Yaqi Duan and Mengdi Wang.
\newblock Minimax-optimal off-policy evaluation with linear function approximation, 2020.
\newblock arXiv:2002.09516.

\bibitem[Dulac-Arnold et~al.(2019)Dulac-Arnold, Mankowitz, and Hester]{dulac2019challenges}
Gabriel Dulac-Arnold, Daniel Mankowitz, and Todd Hester.
\newblock Challenges of real-world reinforcement learning.
\newblock \emph{arXiv preprint arXiv:1904.12901}, 2019.

\bibitem[Ebert et~al.(2021)Ebert, Yang, Schmeckpeper, Bucher, Georgakis, Daniilidis, Finn, and Levine]{ebert2021bridge}
Frederik Ebert, Yanlai Yang, Karl Schmeckpeper, Bernadette Bucher, Georgios Georgakis, Kostas Daniilidis, Chelsea Finn, and Sergey Levine.
\newblock Bridge data: Boosting generalization of robotic skills with cross-domain datasets.
\newblock \emph{arXiv preprint arXiv:2109.13396}, 2021.

\bibitem[Erez et~al.(2015)Erez, Tassa, and Todorov]{erez2015simulation}
Tom Erez, Yuval Tassa, and Emanuel Todorov.
\newblock Simulation tools for model-based robotics: Comparison of bullet, havok, mujoco, ode and physx.
\newblock In \emph{2015 IEEE international conference on robotics and automation (ICRA)}, pp.\  4397--4404. IEEE, 2015.

\bibitem[Esser et~al.(2024)Esser, Kulal, Blattmann, Entezari, Müller, Saini, Levi, Lorenz, Sauer, Boesel, Podell, Dockhorn, English, Lacey, Goodwin, Marek, and Rombach]{esser2024scalingrectifiedflowtransformers}
Patrick Esser, Sumith Kulal, Andreas Blattmann, Rahim Entezari, Jonas Müller, Harry Saini, Yam Levi, Dominik Lorenz, Axel Sauer, Frederic Boesel, Dustin Podell, Tim Dockhorn, Zion English, Kyle Lacey, Alex Goodwin, Yannik Marek, and Robin Rombach.
\newblock Scaling rectified flow transformers for high-resolution image synthesis, 2024.
\newblock URL \url{https://arxiv.org/abs/2403.03206}.

\bibitem[Farajtabar et~al.(2018)Farajtabar, Chow, and Ghavamzadeh]{farajtabar2018more}
Mehrdad Farajtabar, Yinlam Chow, and Mohammad Ghavamzadeh.
\newblock More robust doubly robust off-policy evaluation.
\newblock \emph{arXiv preprint arXiv:1802.03493}, 2018.

\bibitem[Fonteneau et~al.(2013)Fonteneau, Murphy, Wehenkel, and Ernst]{fonteneau13batch}
Raphael Fonteneau, Susan~A. Murphy, Louis Wehenkel, and Damien Ernst.
\newblock Batch mode reinforcement learning based on the synthesis of artificial trajectories.
\newblock \emph{Annals of Operations Research}, 208\penalty0 (1):\penalty0 383--416, 2013.

\bibitem[Fu et~al.(2021)Fu, Norouzi, Nachum, Tucker, Wang, Novikov, Yang, Zhang, Chen, Kumar, et~al.]{fu2021benchmarks}
Justin Fu, Mohammad Norouzi, Ofir Nachum, George Tucker, Ziyu Wang, Alexander Novikov, Mengjiao Yang, Michael~R Zhang, Yutian Chen, Aviral Kumar, et~al.
\newblock Benchmarks for deep off-policy evaluation.
\newblock \emph{arXiv preprint arXiv:2103.16596}, 2021.

\bibitem[Google(2025)]{SharonEtAl2025GeminiImageEditing}
Google.
\newblock Image editing in gemini just got a major upgrade.
\newblock Blog post on “The Keyword”, Google, August~26 2025.
\newblock URL \url{https://blog.google/products/gemini/updated-image-editing-model/}.
\newblock Multimodal Generation Lead, Gemini Apps; Gemini Image Product Lead, Google DeepMind.

\bibitem[Hafner et~al.(2019)Hafner, Lillicrap, Ba, and Norouzi]{hafner2019dream}
Danijar Hafner, Timothy Lillicrap, Jimmy Ba, and Mohammad Norouzi.
\newblock Dream to control: Learning behaviors by latent imagination.
\newblock \emph{arXiv preprint arXiv:1912.01603}, 2019.

\bibitem[Hafner et~al.(2020)Hafner, Lillicrap, Norouzi, and Ba]{hafner2020mastering}
Danijar Hafner, Timothy Lillicrap, Mohammad Norouzi, and Jimmy Ba.
\newblock Mastering atari with discrete world models.
\newblock \emph{arXiv preprint arXiv:2010.02193}, 2020.

\bibitem[He et~al.(2015)He, Zhang, Ren, and Sun]{he2015deepresiduallearningimage}
Kaiming He, Xiangyu Zhang, Shaoqing Ren, and Jian Sun.
\newblock Deep residual learning for image recognition, 2015.
\newblock URL \url{https://arxiv.org/abs/1512.03385}.

\bibitem[Ho et~al.(2022)Ho, Chan, Saharia, Whang, Gao, Gritsenko, Kingma, Poole, Norouzi, Fleet, et~al.]{ho2022imagen}
Jonathan Ho, William Chan, Chitwan Saharia, Jay Whang, Ruiqi Gao, Alexey Gritsenko, Diederik~P Kingma, Ben Poole, Mohammad Norouzi, David~J Fleet, et~al.
\newblock Imagen video: High definition video generation with diffusion models.
\newblock \emph{arXiv preprint arXiv:2210.02303}, 2022.

\bibitem[Hu et~al.(2022)Hu, Shen, Wallis, Allen-Zhu, Li, Wang, Wang, Chen, et~al.]{hu2022lora}
Edward~J Hu, Yelong Shen, Phillip Wallis, Zeyuan Allen-Zhu, Yuanzhi Li, Shean Wang, Lu~Wang, Weizhu Chen, et~al.
\newblock Lora: Low-rank adaptation of large language models.
\newblock \emph{ICLR}, 1\penalty0 (2):\penalty0 3, 2022.

\bibitem[Jiang \& Li(2015)Jiang and Li]{jiang2015doubly}
Nan Jiang and Lihong Li.
\newblock Doubly robust off-policy value evaluation for reinforcement learning.
\newblock \emph{arXiv preprint arXiv:1511.03722}, 2015.

\bibitem[Jiang \& Li(2016)Jiang and Li]{jiang2016doubly}
Nan Jiang and Lihong Li.
\newblock Doubly robust off-policy value evaluation for reinforcement learning.
\newblock In \emph{International conference on machine learning}, pp.\  652--661. PMLR, 2016.

\bibitem[Kaelbling et~al.(1995)Kaelbling, Littman, and Cassandra]{kaelbling1995partially}
Leslie~Pack Kaelbling, Michael~L Littman, and Anthony~R Cassandra.
\newblock Partially observable markov decision processes for artificial intelligence.
\newblock In \emph{International Workshop on Reasoning with Uncertainty in Robotics}, pp.\  146--163. Springer, 1995.

\bibitem[Kaiser et~al.(2019)Kaiser, Babaeizadeh, Milos, Osinski, Campbell, Czechowski, Erhan, Finn, Kozakowski, Levine, et~al.]{kaiser2019model}
Lukasz Kaiser, Mohammad Babaeizadeh, Piotr Milos, Blazej Osinski, Roy~H Campbell, Konrad Czechowski, Dumitru Erhan, Chelsea Finn, Piotr Kozakowski, Sergey Levine, et~al.
\newblock Model-based reinforcement learning for atari.
\newblock \emph{arXiv preprint arXiv:1903.00374}, 2019.

\bibitem[Kallus \& Uehara(2019)Kallus and Uehara]{kallus2019double}
Nathan Kallus and Masatoshi Uehara.
\newblock Double reinforcement learning for efficient off-policy evaluation in {Markov} decision processes.
\newblock \emph{arXiv preprint arXiv:1908.08526}, 2019.

\bibitem[Kanamori et~al.(2009)Kanamori, Hido, and Sugiyama]{kanamori2009least}
Takafumi Kanamori, Shohei Hido, and Masashi Sugiyama.
\newblock A least-squares approach to direct importance estimation.
\newblock \emph{The Journal of Machine Learning Research}, 10:\penalty0 1391--1445, 2009.

\bibitem[Kim et~al.()Kim, Pertsch, Karamcheti, Xiao, Balakrishna, Nair, Rafailov, Foster, Lam, Sanketi, et~al.]{kim2406openvla}
Moo~Jin Kim, Karl Pertsch, Siddharth Karamcheti, Ted Xiao, Ashwin Balakrishna, Suraj Nair, Rafael Rafailov, Ethan Foster, Grace Lam, Pannag Sanketi, et~al.
\newblock Openvla: An open-source vision-language-action model, 2024.
\newblock \emph{URL https://arxiv. org/abs/2406.09246}.

\bibitem[Le et~al.(2019)Le, Voloshin, and Yue]{le2019batch}
Hoang Le, Cameron Voloshin, and Yisong Yue.
\newblock Batch policy learning under constraints.
\newblock In \emph{International Conference on Machine Learning}, pp.\  3703--3712. PMLR, 2019.

\bibitem[Levine et~al.(2020)Levine, Kumar, Tucker, and Fu]{levine2020offline}
Sergey Levine, Aviral Kumar, George Tucker, and Justin Fu.
\newblock Offline reinforcement learning: Tutorial, review, and perspectives on open problems.
\newblock \emph{arXiv preprint arXiv:2005.01643}, 2020.

\bibitem[Li et~al.(2024)Li, Hsu, Gu, Pertsch, Mees, Walke, Fu, Lunawat, Sieh, Kirmani, et~al.]{li2024evaluating}
Xuanlin Li, Kyle Hsu, Jiayuan Gu, Karl Pertsch, Oier Mees, Homer~Rich Walke, Chuyuan Fu, Ishikaa Lunawat, Isabel Sieh, Sean Kirmani, et~al.
\newblock Evaluating real-world robot manipulation policies in simulation.
\newblock \emph{arXiv preprint arXiv:2405.05941}, 2024.

\bibitem[Li et~al.(2025)Li, Zhu, Wen, Shen, and Xu]{li2025worldeval}
Yaxuan Li, Yichen Zhu, Junjie Wen, Chaomin Shen, and Yi~Xu.
\newblock Worldeval: World model as real-world robot policies evaluator.
\newblock \emph{arXiv preprint arXiv:2505.19017}, 2025.

\bibitem[Liu et~al.(2024)Liu, Orru, Vakil, Paxton, Shafiullah, and Pinto]{liu2024ok}
Peiqi Liu, Yaswanth Orru, Jay Vakil, Chris Paxton, Nur Muhammad~Mahi Shafiullah, and Lerrel Pinto.
\newblock Ok-robot: What really matters in integrating open-knowledge models for robotics.
\newblock \emph{arXiv preprint arXiv:2401.12202}, 2024.

\bibitem[Ma et~al.(2025)Ma, Wang, Chen, Jia, Liu, Li, Chen, and Qiao]{ma2025lattelatentdiffusiontransformer}
Xin Ma, Yaohui Wang, Xinyuan Chen, Gengyun Jia, Ziwei Liu, Yuan-Fang Li, Cunjian Chen, and Yu~Qiao.
\newblock Latte: Latent diffusion transformer for video generation, 2025.
\newblock URL \url{https://arxiv.org/abs/2401.03048}.

\bibitem[Munos et~al.(2016)Munos, Stepleton, Harutyunyan, and Bellemare]{munos2016safe}
R.~Munos, T.~Stepleton, A.~Harutyunyan, and M.~Bellemare.
\newblock Safe and efficient off-policy reinforcement learning.
\newblock In \emph{Advances in Neural Information Processing Systems}, pp.\  1054--1062, 2016.

\bibitem[Nachum et~al.(2019)Nachum, Chow, Dai, and Li]{nachum2019dualdice}
Ofir Nachum, Yinlam Chow, Bo~Dai, and Lihong Li.
\newblock Dualdice: Behavior-agnostic estimation of discounted stationary distribution corrections.
\newblock \emph{Advances in neural information processing systems}, 32, 2019.

\bibitem[Nguyen et~al.(2010)Nguyen, Wainwright, and Jordan]{nguyen2010estimating}
XuanLong Nguyen, Martin~J Wainwright, and Michael~I Jordan.
\newblock Estimating divergence functionals and the likelihood ratio by convex risk minimization.
\newblock \emph{IEEE Transactions on Information Theory}, 56\penalty0 (11):\penalty0 5847--5861, 2010.

\bibitem[NVIDIA et~al.(2025)NVIDIA, :, Agarwal, Ali, Bala, Balaji, Barker, Cai, Chattopadhyay, Chen, Cui, Ding, Dworakowski, Fan, Fenzi, Ferroni, Fidler, Fox, Ge, Ge, Gu, Gururani, He, Huang, Huffman, Jannaty, Jin, Kim, Klár, Lam, Lan, Leal-Taixe, Li, Li, Lin, Lin, Ling, Liu, Liu, Luo, Ma, Mao, Mo, Mousavian, Nah, Niverty, Page, Paschalidou, Patel, Pavao, Ramezanali, Reda, Ren, Sabavat, Schmerling, Shi, Stefaniak, Tang, Tchapmi, Tredak, Tseng, Varghese, Wang, Wang, Wang, Wang, Wei, Wei, Wu, Xu, Yang, Yen-Chen, Zeng, Zeng, Zhang, Zhang, Zhang, Zhao, and Zolkowski]{nvidia2025cosmosworldfoundationmodel}
NVIDIA, :, Niket Agarwal, Arslan Ali, Maciej Bala, Yogesh Balaji, Erik Barker, Tiffany Cai, Prithvijit Chattopadhyay, Yongxin Chen, Yin Cui, Yifan Ding, Daniel Dworakowski, Jiaojiao Fan, Michele Fenzi, Francesco Ferroni, Sanja Fidler, Dieter Fox, Songwei Ge, Yunhao Ge, Jinwei Gu, Siddharth Gururani, Ethan He, Jiahui Huang, Jacob Huffman, Pooya Jannaty, Jingyi Jin, Seung~Wook Kim, Gergely Klár, Grace Lam, Shiyi Lan, Laura Leal-Taixe, Anqi Li, Zhaoshuo Li, Chen-Hsuan Lin, Tsung-Yi Lin, Huan Ling, Ming-Yu Liu, Xian Liu, Alice Luo, Qianli Ma, Hanzi Mao, Kaichun Mo, Arsalan Mousavian, Seungjun Nah, Sriharsha Niverty, David Page, Despoina Paschalidou, Zeeshan Patel, Lindsey Pavao, Morteza Ramezanali, Fitsum Reda, Xiaowei Ren, Vasanth Rao~Naik Sabavat, Ed~Schmerling, Stella Shi, Bartosz Stefaniak, Shitao Tang, Lyne Tchapmi, Przemek Tredak, Wei-Cheng Tseng, Jibin Varghese, Hao Wang, Haoxiang Wang, Heng Wang, Ting-Chun Wang, Fangyin Wei, Xinyue Wei, Jay~Zhangjie Wu, Jiashu Xu, Wei Yang, Lin Yen-Chen, Xiaohui Zeng,
  Yu~Zeng, Jing Zhang, Qinsheng Zhang, Yuxuan Zhang, Qingqing Zhao, and Artur Zolkowski.
\newblock Cosmos world foundation model platform for physical ai, 2025.
\newblock URL \url{https://arxiv.org/abs/2501.03575}.

\bibitem[{Octo Model Team} et~al.(2024){Octo Model Team}, Ghosh, Walke, Pertsch, Black, Mees, Dasari, Hejna, Xu, Luo, Kreiman, Tan, Chen, Sanketi, Vuong, Xiao, Sadigh, Finn, and Levine]{octo_2023}
{Octo Model Team}, Dibya Ghosh, Homer Walke, Karl Pertsch, Kevin Black, Oier Mees, Sudeep Dasari, Joey Hejna, Charles Xu, Jianlan Luo, Tobias Kreiman, {You Liang} Tan, Lawrence~Yunliang Chen, Pannag Sanketi, Quan Vuong, Ted Xiao, Dorsa Sadigh, Chelsea Finn, and Sergey Levine.
\newblock Octo: An open-source generalist robot policy.
\newblock In \emph{Proceedings of Robotics: Science and Systems}, Delft, Netherlands, 2024.

\bibitem[O'Neill et~al.(2023)O'Neill, Rehman, Gupta, Maddukuri, Gupta, Padalkar, Lee, Pooley, Gupta, Mandlekar, et~al.]{o2023open}
Abby O'Neill, Abdul Rehman, Abhinav Gupta, Abhiram Maddukuri, Abhishek Gupta, Abhishek Padalkar, Abraham Lee, Acorn Pooley, Agrim Gupta, Ajay Mandlekar, et~al.
\newblock Open x-embodiment: Robotic learning datasets and rt-x models.
\newblock \emph{arXiv preprint arXiv:2310.08864}, 2023.

\bibitem[OpenAI et~al.(2024)OpenAI, :, Hurst, Lerer, Goucher, Perelman, Ramesh, Clark, Ostrow, Welihinda, Hayes, Radford, Mądry, Baker-Whitcomb, Beutel, Borzunov, Carney, Chow, Kirillov, Nichol, Paino, Renzin, Passos, Kirillov, Christakis, Conneau, Kamali, Jabri, Moyer, Tam, Crookes, Tootoochian, Tootoonchian, Kumar, Vallone, Karpathy, Braunstein, Cann, Codispoti, Galu, Kondrich, Tulloch, Mishchenko, Baek, Jiang, Pelisse, Woodford, Gosalia, Dhar, Pantuliano, Nayak, Oliver, Zoph, Ghorbani, Leimberger, Rossen, Sokolowsky, Wang, Zweig, Hoover, Samic, McGrew, Spero, Giertler, Cheng, Lightcap, Walkin, Quinn, Guarraci, Hsu, Kellogg, Eastman, Lugaresi, Wainwright, Bassin, Hudson, Chu, Nelson, Li, Shern, Conger, Barette, Voss, Ding, Lu, Zhang, Beaumont, Hallacy, Koch, Gibson, Kim, Choi, McLeavey, Hesse, Fischer, Winter, Czarnecki, Jarvis, Wei, Koumouzelis, Sherburn, Kappler, Levin, Levy, Carr, Farhi, Mely, Robinson, Sasaki, Jin, Valladares, Tsipras, Li, Nguyen, Findlay, Oiwoh, Wong, Asdar, Proehl, Yang, Antonow,
  Kramer, Peterson, Sigler, Wallace, Brevdo, Mays, Khorasani, Such, Raso, Zhang, von Lohmann, Sulit, Goh, Oden, Salmon, Starace, Brockman, Salman, Bao, Hu, Wong, Wang, Schmidt, Whitney, Jun, Kirchner, de~Oliveira~Pinto, Ren, Chang, Chung, Kivlichan, O'Connell, O'Connell, Osband, Silber, Sohl, Okuyucu, Lan, Kostrikov, Sutskever, Kanitscheider, Gulrajani, Coxon, Menick, Pachocki, Aung, Betker, Crooks, Lennon, Kiros, Leike, Park, Kwon, Phang, Teplitz, Wei, Wolfe, Chen, Harris, Varavva, Lee, Shieh, Lin, Yu, Weng, Tang, Yu, Jang, Candela, Beutler, Landers, Parish, Heidecke, Schulman, Lachman, McKay, Uesato, Ward, Kim, Huizinga, Sitkin, Kraaijeveld, Gross, Kaplan, Snyder, Achiam, Jiao, Lee, Zhuang, Harriman, Fricke, Hayashi, Singhal, Shi, Karthik, Wood, Rimbach, Hsu, Nguyen, Gu-Lemberg, Button, Liu, Howe, Muthukumar, Luther, Ahmad, Kai, Itow, Workman, Pathak, Chen, Jing, Guy, Fedus, Zhou, Mamitsuka, Weng, McCallum, Held, Ouyang, Feuvrier, Zhang, Kondraciuk, Kaiser, Hewitt, Metz, Doshi, Aflak, Simens, Boyd,
  Thompson, Dukhan, Chen, Gray, Hudnall, Zhang, Aljubeh, Litwin, Zeng, Johnson, Shetty, Gupta, Shah, Yatbaz, Yang, Zhong, Glaese, Chen, Janner, Lampe, Petrov, Wu, Wang, Fradin, Pokrass, Castro, de~Castro, Pavlov, Brundage, Wang, Khan, Murati, Bavarian, Lin, Yesildal, Soto, Gimelshein, Cone, Staudacher, Summers, LaFontaine, Chowdhury, Ryder, Stathas, Turley, Tezak, Felix, Kudige, Keskar, Deutsch, Bundick, Puckett, Nachum, Okelola, Boiko, Murk, Jaffe, Watkins, Godement, Campbell-Moore, Chao, McMillan, Belov, Su, Bak, Bakkum, Deng, Dolan, Hoeschele, Welinder, Tillet, Pronin, Tillet, Dhariwal, Yuan, Dias, Lim, Arora, Troll, Lin, Lopes, Puri, Miyara, Leike, Gaubert, Zamani, Wang, Donnelly, Honsby, Smith, Sahai, Ramchandani, Huet, Carmichael, Zellers, Chen, Chen, Nigmatullin, Cheu, Jain, Altman, Schoenholz, Toizer, Miserendino, Agarwal, Culver, Ethersmith, Gray, Grove, Metzger, Hermani, Jain, Zhao, Wu, Jomoto, Wu, Shuaiqi, Xia, Phene, Papay, Narayanan, Coffey, Lee, Hall, Balaji, Broda, Stramer, Xu, Gogineni,
  Christianson, Sanders, Patwardhan, Cunninghman, Degry, Dimson, Raoux, Shadwell, Zheng, Underwood, Markov, Sherbakov, Rubin, Stasi, Kaftan, Heywood, Peterson, Walters, Eloundou, Qi, Moeller, Monaco, Kuo, Fomenko, Chang, Zheng, Zhou, Manassra, Sheu, Zaremba, Patil, Qian, Kim, Cheng, Zhang, He, Zhang, Jin, Dai, and Malkov]{openai2024gpt4ocard}
OpenAI, :, Aaron Hurst, Adam Lerer, Adam~P. Goucher, Adam Perelman, Aditya Ramesh, Aidan Clark, AJ~Ostrow, Akila Welihinda, Alan Hayes, Alec Radford, Aleksander Mądry, Alex Baker-Whitcomb, Alex Beutel, Alex Borzunov, Alex Carney, Alex Chow, Alex Kirillov, Alex Nichol, Alex Paino, Alex Renzin, Alex~Tachard Passos, Alexander Kirillov, Alexi Christakis, Alexis Conneau, Ali Kamali, Allan Jabri, Allison Moyer, Allison Tam, Amadou Crookes, Amin Tootoochian, Amin Tootoonchian, Ananya Kumar, Andrea Vallone, Andrej Karpathy, Andrew Braunstein, Andrew Cann, Andrew Codispoti, Andrew Galu, Andrew Kondrich, Andrew Tulloch, Andrey Mishchenko, Angela Baek, Angela Jiang, Antoine Pelisse, Antonia Woodford, Anuj Gosalia, Arka Dhar, Ashley Pantuliano, Avi Nayak, Avital Oliver, Barret Zoph, Behrooz Ghorbani, Ben Leimberger, Ben Rossen, Ben Sokolowsky, Ben Wang, Benjamin Zweig, Beth Hoover, Blake Samic, Bob McGrew, Bobby Spero, Bogo Giertler, Bowen Cheng, Brad Lightcap, Brandon Walkin, Brendan Quinn, Brian Guarraci, Brian Hsu,
  Bright Kellogg, Brydon Eastman, Camillo Lugaresi, Carroll Wainwright, Cary Bassin, Cary Hudson, Casey Chu, Chad Nelson, Chak Li, Chan~Jun Shern, Channing Conger, Charlotte Barette, Chelsea Voss, Chen Ding, Cheng Lu, Chong Zhang, Chris Beaumont, Chris Hallacy, Chris Koch, Christian Gibson, Christina Kim, Christine Choi, Christine McLeavey, Christopher Hesse, Claudia Fischer, Clemens Winter, Coley Czarnecki, Colin Jarvis, Colin Wei, Constantin Koumouzelis, Dane Sherburn, Daniel Kappler, Daniel Levin, Daniel Levy, David Carr, David Farhi, David Mely, David Robinson, David Sasaki, Denny Jin, Dev Valladares, Dimitris Tsipras, Doug Li, Duc~Phong Nguyen, Duncan Findlay, Edede Oiwoh, Edmund Wong, Ehsan Asdar, Elizabeth Proehl, Elizabeth Yang, Eric Antonow, Eric Kramer, Eric Peterson, Eric Sigler, Eric Wallace, Eugene Brevdo, Evan Mays, Farzad Khorasani, Felipe~Petroski Such, Filippo Raso, Francis Zhang, Fred von Lohmann, Freddie Sulit, Gabriel Goh, Gene Oden, Geoff Salmon, Giulio Starace, Greg Brockman, Hadi
  Salman, Haiming Bao, Haitang Hu, Hannah Wong, Haoyu Wang, Heather Schmidt, Heather Whitney, Heewoo Jun, Hendrik Kirchner, Henrique~Ponde de~Oliveira~Pinto, Hongyu Ren, Huiwen Chang, Hyung~Won Chung, Ian Kivlichan, Ian O'Connell, Ian O'Connell, Ian Osband, Ian Silber, Ian Sohl, Ibrahim Okuyucu, Ikai Lan, Ilya Kostrikov, Ilya Sutskever, Ingmar Kanitscheider, Ishaan Gulrajani, Jacob Coxon, Jacob Menick, Jakub Pachocki, James Aung, James Betker, James Crooks, James Lennon, Jamie Kiros, Jan Leike, Jane Park, Jason Kwon, Jason Phang, Jason Teplitz, Jason Wei, Jason Wolfe, Jay Chen, Jeff Harris, Jenia Varavva, Jessica~Gan Lee, Jessica Shieh, Ji~Lin, Jiahui Yu, Jiayi Weng, Jie Tang, Jieqi Yu, Joanne Jang, Joaquin~Quinonero Candela, Joe Beutler, Joe Landers, Joel Parish, Johannes Heidecke, John Schulman, Jonathan Lachman, Jonathan McKay, Jonathan Uesato, Jonathan Ward, Jong~Wook Kim, Joost Huizinga, Jordan Sitkin, Jos Kraaijeveld, Josh Gross, Josh Kaplan, Josh Snyder, Joshua Achiam, Joy Jiao, Joyce Lee, Juntang
  Zhuang, Justyn Harriman, Kai Fricke, Kai Hayashi, Karan Singhal, Katy Shi, Kavin Karthik, Kayla Wood, Kendra Rimbach, Kenny Hsu, Kenny Nguyen, Keren Gu-Lemberg, Kevin Button, Kevin Liu, Kiel Howe, Krithika Muthukumar, Kyle Luther, Lama Ahmad, Larry Kai, Lauren Itow, Lauren Workman, Leher Pathak, Leo Chen, Li~Jing, Lia Guy, Liam Fedus, Liang Zhou, Lien Mamitsuka, Lilian Weng, Lindsay McCallum, Lindsey Held, Long Ouyang, Louis Feuvrier, Lu~Zhang, Lukas Kondraciuk, Lukasz Kaiser, Luke Hewitt, Luke Metz, Lyric Doshi, Mada Aflak, Maddie Simens, Madelaine Boyd, Madeleine Thompson, Marat Dukhan, Mark Chen, Mark Gray, Mark Hudnall, Marvin Zhang, Marwan Aljubeh, Mateusz Litwin, Matthew Zeng, Max Johnson, Maya Shetty, Mayank Gupta, Meghan Shah, Mehmet Yatbaz, Meng~Jia Yang, Mengchao Zhong, Mia Glaese, Mianna Chen, Michael Janner, Michael Lampe, Michael Petrov, Michael Wu, Michele Wang, Michelle Fradin, Michelle Pokrass, Miguel Castro, Miguel Oom~Temudo de~Castro, Mikhail Pavlov, Miles Brundage, Miles Wang, Minal
  Khan, Mira Murati, Mo~Bavarian, Molly Lin, Murat Yesildal, Nacho Soto, Natalia Gimelshein, Natalie Cone, Natalie Staudacher, Natalie Summers, Natan LaFontaine, Neil Chowdhury, Nick Ryder, Nick Stathas, Nick Turley, Nik Tezak, Niko Felix, Nithanth Kudige, Nitish Keskar, Noah Deutsch, Noel Bundick, Nora Puckett, Ofir Nachum, Ola Okelola, Oleg Boiko, Oleg Murk, Oliver Jaffe, Olivia Watkins, Olivier Godement, Owen Campbell-Moore, Patrick Chao, Paul McMillan, Pavel Belov, Peng Su, Peter Bak, Peter Bakkum, Peter Deng, Peter Dolan, Peter Hoeschele, Peter Welinder, Phil Tillet, Philip Pronin, Philippe Tillet, Prafulla Dhariwal, Qiming Yuan, Rachel Dias, Rachel Lim, Rahul Arora, Rajan Troll, Randall Lin, Rapha~Gontijo Lopes, Raul Puri, Reah Miyara, Reimar Leike, Renaud Gaubert, Reza Zamani, Ricky Wang, Rob Donnelly, Rob Honsby, Rocky Smith, Rohan Sahai, Rohit Ramchandani, Romain Huet, Rory Carmichael, Rowan Zellers, Roy Chen, Ruby Chen, Ruslan Nigmatullin, Ryan Cheu, Saachi Jain, Sam Altman, Sam Schoenholz, Sam
  Toizer, Samuel Miserendino, Sandhini Agarwal, Sara Culver, Scott Ethersmith, Scott Gray, Sean Grove, Sean Metzger, Shamez Hermani, Shantanu Jain, Shengjia Zhao, Sherwin Wu, Shino Jomoto, Shirong Wu, Shuaiqi, Xia, Sonia Phene, Spencer Papay, Srinivas Narayanan, Steve Coffey, Steve Lee, Stewart Hall, Suchir Balaji, Tal Broda, Tal Stramer, Tao Xu, Tarun Gogineni, Taya Christianson, Ted Sanders, Tejal Patwardhan, Thomas Cunninghman, Thomas Degry, Thomas Dimson, Thomas Raoux, Thomas Shadwell, Tianhao Zheng, Todd Underwood, Todor Markov, Toki Sherbakov, Tom Rubin, Tom Stasi, Tomer Kaftan, Tristan Heywood, Troy Peterson, Tyce Walters, Tyna Eloundou, Valerie Qi, Veit Moeller, Vinnie Monaco, Vishal Kuo, Vlad Fomenko, Wayne Chang, Weiyi Zheng, Wenda Zhou, Wesam Manassra, Will Sheu, Wojciech Zaremba, Yash Patil, Yilei Qian, Yongjik Kim, Youlong Cheng, Yu~Zhang, Yuchen He, Yuchen Zhang, Yujia Jin, Yunxing Dai, and Yury Malkov.
\newblock Gpt-4o system card, 2024.
\newblock URL \url{https://arxiv.org/abs/2410.21276}.

\bibitem[Peebles \& Xie(2023)Peebles and Xie]{peebles2023scalable}
William Peebles and Saining Xie.
\newblock Scalable diffusion models with transformers.
\newblock In \emph{Proceedings of the IEEE/CVF International Conference on Computer Vision}, pp.\  4195--4205, 2023.

\bibitem[Precup et~al.(2000)Precup, Sutton, and Singh]{Precup00ET}
Doina Precup, Richard~S. Sutton, and Satinder~P. Singh.
\newblock Eligibility traces for off-policy policy evaluation.
\newblock In \emph{Proceedings of the 17th International Conference on Machine Learning}, pp.\  759--766, 2000.

\bibitem[Puterman(2014)]{puterman2014markov}
Martin~L Puterman.
\newblock \emph{Markov decision processes: discrete stochastic dynamic programming}.
\newblock John Wiley \& Sons, 2014.

\bibitem[Salvato et~al.(2021)Salvato, Fenu, Medvet, and Pellegrino]{salvato2021crossing}
Erica Salvato, Gianfranco Fenu, Eric Medvet, and Felice~Andrea Pellegrino.
\newblock Crossing the reality gap: A survey on sim-to-real transferability of robot controllers in reinforcement learning.
\newblock \emph{IEEE Access}, 9:\penalty0 153171--153187, 2021.

\bibitem[Shafiullah et~al.(2023)Shafiullah, Rai, Etukuru, Liu, Misra, Chintala, and Pinto]{shafiullah2023bringing}
Nur Muhammad~Mahi Shafiullah, Anant Rai, Haritheja Etukuru, Yiqian Liu, Ishan Misra, Soumith Chintala, and Lerrel Pinto.
\newblock On bringing robots home.
\newblock \emph{arXiv preprint arXiv:2311.16098}, 2023.

\bibitem[Singer et~al.(2022)Singer, Polyak, Hayes, Yin, An, Zhang, Hu, Yang, Ashual, Gafni, et~al.]{singer2022make}
Uriel Singer, Adam Polyak, Thomas Hayes, Xi~Yin, Jie An, Songyang Zhang, Qiyuan Hu, Harry Yang, Oron Ashual, Oran Gafni, et~al.
\newblock Make-a-video: Text-to-video generation without text-video data.
\newblock \emph{arXiv preprint arXiv:2209.14792}, 2022.

\bibitem[Soljacic et~al.(2024)Soljacic, Law, Chita-Tegmark, and Scheutz]{soljacic2024robots}
Fran Soljacic, Theresa Law, Meia Chita-Tegmark, and Matthias Scheutz.
\newblock Robots in healthcare as envisioned by care professionals.
\newblock \emph{Intelligent Service Robotics}, pp.\  1--17, 2024.

\bibitem[S{\"u}nderhauf et~al.(2018)S{\"u}nderhauf, Brock, Scheirer, Hadsell, Fox, Leitner, Upcroft, Abbeel, Burgard, Milford, et~al.]{sunderhauf2018limits}
Niko S{\"u}nderhauf, Oliver Brock, Walter Scheirer, Raia Hadsell, Dieter Fox, J{\"u}rgen Leitner, Ben Upcroft, Pieter Abbeel, Wolfram Burgard, Michael Milford, et~al.
\newblock The limits and potentials of deep learning for robotics.
\newblock \emph{The International journal of robotics research}, 37\penalty0 (4-5):\penalty0 405--420, 2018.

\bibitem[Sutton et~al.(2009)Sutton, Maei, Precup, Bhatnagar, Silver, Szepesv{\'a}ri, and Wiewiora]{sutton2009fast}
Richard~S Sutton, Hamid~Reza Maei, Doina Precup, Shalabh Bhatnagar, David Silver, Csaba Szepesv{\'a}ri, and Eric Wiewiora.
\newblock Fast gradient-descent methods for temporal-difference learning with linear function approximation.
\newblock In \emph{Proceedings of the 26th annual international conference on machine learning}, pp.\  993--1000, 2009.

\bibitem[Sutton et~al.(2016)Sutton, Mahmood, and White]{sutton2016emphatic}
Richard~S Sutton, A~Rupam Mahmood, and Martha White.
\newblock An emphatic approach to the problem of off-policy temporal-difference learning.
\newblock \emph{Journal of Machine Learning Research}, 17\penalty0 (73):\penalty0 1--29, 2016.

\bibitem[Tedrake et~al.(2019)]{tedrake2019drake}
Russ Tedrake et~al.
\newblock Drake: Model-based design and verification for robotics.
\newblock 2019.

\bibitem[Thomas et~al.(2015{\natexlab{a}})Thomas, Theocharous, and Ghavamzadeh]{Thomas15HCPE}
P.~Thomas, G.~Theocharous, and M.~Ghavamzadeh.
\newblock High confidence off-policy evaluation.
\newblock In \emph{Proceedings of the 29th Conference on Artificial Intelligence}, 2015{\natexlab{a}}.

\bibitem[Thomas \& Brunskill(2016)Thomas and Brunskill]{thomas2016data}
Philip Thomas and Emma Brunskill.
\newblock Data-efficient off-policy policy evaluation for reinforcement learning.
\newblock In \emph{International Conference on Machine Learning}, pp.\  2139--2148. PMLR, 2016.

\bibitem[Thomas et~al.(2015{\natexlab{b}})Thomas, Theocharous, and Ghavamzadeh]{thomas2015high}
Philip Thomas, Georgios Theocharous, and Mohammad Ghavamzadeh.
\newblock High-confidence off-policy evaluation.
\newblock In \emph{Proceedings of the AAAI Conference on Artificial Intelligence}, volume~29, 2015{\natexlab{b}}.

\bibitem[Todorov et~al.(2012)Todorov, Erez, and Tassa]{todorov2012mujoco}
Emanuel Todorov, Tom Erez, and Yuval Tassa.
\newblock Mujoco: A physics engine for model-based control.
\newblock In \emph{2012 IEEE/RSJ international conference on intelligent robots and systems}, pp.\  5026--5033. IEEE, 2012.

\bibitem[Valevski et~al.(2024)Valevski, Leviathan, Arar, and Fruchter]{valevski2024diffusion}
Dani Valevski, Yaniv Leviathan, Moab Arar, and Shlomi Fruchter.
\newblock Diffusion models are real-time game engines.
\newblock \emph{arXiv preprint arXiv:2408.14837}, 2024.

\bibitem[Villegas et~al.(2022)Villegas, Babaeizadeh, Kindermans, Moraldo, Zhang, Saffar, Castro, Kunze, and Erhan]{villegas2022phenaki}
Ruben Villegas, Mohammad Babaeizadeh, Pieter-Jan Kindermans, Hernan Moraldo, Han Zhang, Mohammad~Taghi Saffar, Santiago Castro, Julius Kunze, and Dumitru Erhan.
\newblock Phenaki: Variable length video generation from open domain textual description.
\newblock \emph{arXiv preprint arXiv:2210.02399}, 2022.

\bibitem[Walke et~al.(2023)Walke, Black, Zhao, Vuong, Zheng, Hansen-Estruch, He, Myers, Kim, Du, et~al.]{walke2023bridgedata}
Homer~Rich Walke, Kevin Black, Tony~Z Zhao, Quan Vuong, Chongyi Zheng, Philippe Hansen-Estruch, Andre~Wang He, Vivek Myers, Moo~Jin Kim, Max Du, et~al.
\newblock Bridgedata v2: A dataset for robot learning at scale.
\newblock In \emph{Conference on Robot Learning}, pp.\  1723--1736. PMLR, 2023.

\bibitem[Wang et~al.(2024)Wang, Bai, Tan, Wang, Fan, Bai, Chen, Liu, Wang, Ge, Fan, Dang, Du, Ren, Men, Liu, Zhou, Zhou, and Lin]{wang2024qwen2vlenhancingvisionlanguagemodels}
Peng Wang, Shuai Bai, Sinan Tan, Shijie Wang, Zhihao Fan, Jinze Bai, Keqin Chen, Xuejing Liu, Jialin Wang, Wenbin Ge, Yang Fan, Kai Dang, Mengfei Du, Xuancheng Ren, Rui Men, Dayiheng Liu, Chang Zhou, Jingren Zhou, and Junyang Lin.
\newblock Qwen2-vl: Enhancing vision-language model's perception of the world at any resolution, 2024.
\newblock URL \url{https://arxiv.org/abs/2409.12191}.

\bibitem[Wen et~al.(2025)Wen, Zhu, Li, Tang, Shen, and Feng]{wen2025dexvlavisionlanguagemodelplugin}
Junjie Wen, Yichen Zhu, Jinming Li, Zhibin Tang, Chaomin Shen, and Feifei Feng.
\newblock Dexvla: Vision-language model with plug-in diffusion expert for general robot control, 2025.
\newblock URL \url{https://arxiv.org/abs/2502.05855}.

\bibitem[Xiao et~al.(2019)Xiao, Wu, Ma, Schuurmans, and M{\"u}ller]{xiao2019learning}
Chenjun Xiao, Yifan Wu, Chen Ma, Dale Schuurmans, and Martin M{\"u}ller.
\newblock Learning to combat compounding-error in model-based reinforcement learning.
\newblock \emph{arXiv preprint arXiv:1912.11206}, 2019.

\bibitem[Yang et~al.(2020)Yang, Nachum, Dai, Li, and Schuurmans]{yang2020off}
Mengjiao Yang, Ofir Nachum, Bo~Dai, Lihong Li, and Dale Schuurmans.
\newblock Off-policy evaluation via the regularized lagrangian.
\newblock \emph{Advances in Neural Information Processing Systems}, 33:\penalty0 6551--6561, 2020.

\bibitem[Yang et~al.(2023)Yang, Du, Ghasemipour, Tompson, Schuurmans, and Abbeel]{yang2023learning}
Mengjiao Yang, Yilun Du, Kamyar Ghasemipour, Jonathan Tompson, Dale Schuurmans, and Pieter Abbeel.
\newblock Learning interactive real-world simulators.
\newblock \emph{arXiv preprint arXiv:2310.06114}, 2023.

\bibitem[Yu et~al.(2020)Yu, Thomas, Yu, Ermon, Zou, Levine, Finn, and Ma]{yu2020mopo}
Tianhe Yu, Garrett Thomas, Lantao Yu, Stefano Ermon, James~Y Zou, Sergey Levine, Chelsea Finn, and Tengyu Ma.
\newblock Mopo: Model-based offline policy optimization.
\newblock \emph{Advances in Neural Information Processing Systems}, 33:\penalty0 14129--14142, 2020.

\bibitem[Zhang et~al.(2021)Zhang, Paine, Nachum, Paduraru, Tucker, Wang, and Norouzi]{zhang2021autoregressive}
Michael~R Zhang, Tom~Le Paine, Ofir Nachum, Cosmin Paduraru, George Tucker, Ziyu Wang, and Mohammad Norouzi.
\newblock Autoregressive dynamics models for offline policy evaluation and optimization.
\newblock \emph{arXiv preprint arXiv:2104.13877}, 2021.

\bibitem[Zhao et~al.(2020)Zhao, Queralta, and Westerlund]{zhao2020sim}
Wenshuai Zhao, Jorge~Pe{\~n}a Queralta, and Tomi Westerlund.
\newblock Sim-to-real transfer in deep reinforcement learning for robotics: a survey.
\newblock In \emph{2020 IEEE symposium series on computational intelligence (SSCI)}, pp.\  737--744. IEEE, 2020.

\end{thebibliography}
